\documentclass[lettersize,journal]{IEEEtran}
\usepackage{amsmath, amsfonts, amsthm} 
\usepackage{algorithmic}
\usepackage{algorithm}
\usepackage{array}
\usepackage[caption=false,font=normalsize,labelfont=sf,textfont=sf]{subfig}
\usepackage{textcomp}
\usepackage{stfloats}
\usepackage{url}
\usepackage{verbatim}
\usepackage{graphicx}
\usepackage{cite}
\usepackage{multirow}
\newcolumntype{P}[1]{>{\centering\arraybackslash}p{#1}}

\usepackage[colorlinks=true]{hyperref}
\newcommand{\theoremheader}[1]{\noindent\textbf{#1}\quad}

\usepackage{booktabs} 

\newtheorem{remark}{Remark}

\begin{document}
	
\title{Active Contour Models Driven by Hyperbolic Mean Curvature Flow for Image Segmentation}

\author{
\IEEEauthorblockN{
Saiyu Hu,
Chunlei He,
Jianfeng Zhang,
Dexing Kong\textsuperscript{*},
and Shoujun Huang\textsuperscript{*}
}\\[2mm]
\IEEEauthorblockA{
\textit{College of Mathematical Medicine, Zhejiang Normal University}\\
Jinhua 321004, China\\
}

\thanks{*Corresponding authors: dkong@zjnu.edu.cn (D. X. Kong); sjhuang@zjnu.edu.cn (S. J. Huang)} 
\thanks{This work was supported by Zhejiang Normal University (Grant Nos. YS304222929, YS304222977) and National Natural Science Foundation of China (Grant Nos. 12301676, 12090020, 12090025).}
}

\maketitle

	\begin{abstract} 

Parabolic mean curvature flow-driven active contour models (PMCF-ACMs) are widely used for image segmentation, yet they suffer severe degradation under high-intensity noise because gradient-descent evolutions exhibit the well-known zig-zag phenomenon. To overcome this drawback, we propose hyperbolic mean curvature flow-driven ACMs (HMCF-ACMs). This novel framework incorporates an adjustable acceleration field to autonomously regulate curve evolution smoothness, providing dual degrees of freedom for adaptive selection of both initial contours and velocity fields. We rigorously prove that HMCF-ACMs are normal flows and establish their numerical equivalence to wave equations through a level set formulation with signed distance functions. An efficient numerical scheme combining spectral discretization and optimized temporal integration is developed to solve the governing equations, and its stability condition is derived through Fourier analysis. Extensive experiments on natural and medical images validate that HMCF-ACMs achieve superior performance under high-noise conditions, demonstrating reduced parameter sensitivity, enhanced noise robustness, and improved segmentation accuracy compared to PMCF-ACMs.

	\end{abstract}
	
	
	\section{Introduction}
	\IEEEPARstart{I}{mage} segmentation is a fundamental technique in computer vision and image processing, which is essential for partitioning digital images into semantically meaningful regions of interest. This process enables precise delineation of object boundaries and supports subsequent quantitative analysis and decision-making. Its applications span a wide range of fields, including medical image analysis \cite{moccia2018blood}, autonomous driving \cite{yao2023radar}, defect detection \cite{luo2020automated}, and remote sensing image analysis \cite{soudagar2025enhanced}.
	
	Current segmentation methodologies are broadly classified into two principal categories: image feature-driven approaches and prior knowledge-constrained techniques, as summarized in~\cite{hu2016automatic}. The first category integrates classical methodologies including region growing algorithms \cite{li2024infrared}, level set methods \cite{chan2001active,vese2002multiphase,li2011level,ding2018active,niaz2023edge}, threshold-based segmentation \cite{zhao2021chaotic}, and graph-theoretical formulations \cite{boykov2001interactive}. The latter category encompasses parametric probability density-based active contour models (ACMs) utilizing either Maximum Likelihood Estimation \cite{rahmati2012mammography} or Maximum a Posteriori Estimation \cite{wang2020active}, along with non-parametric counterparts grounded in statistical theory~\cite{andreetto2007non} or information-theoretic principles~\cite{kim2005nonparametric}. Among these, the ACM pioneered by Kass et al. \cite{kass1988snakes} established a seminal paradigm for contour evolution through energy functional optimization. Subsequent developments in ACM implementations have principally evolved along two distinct methodological trajectories: edge-driven \cite{xu1997gradient,caselles1997geodesic,li2010distance,li2005level} and region-based \cite{mumford1989optimal,chan2001active,li2011level,ding2018active,niaz2023edge} models.	

	Edge-based ACMs evolve contours by tracking image gradients and low-level edge features. They excel for images with sharp edges but struggle with blurred or weak edges, often requiring manual initialization and parameter tuning. Although advanced variants like Gradient Vector Flow \cite{xu1997gradient}, Geodesic Active Contours (GAC) \cite{caselles1997geodesic}, and subsequent methods \cite{caselles1997geodesic,li2010distance,li2005level} enhance capture range and reduce manual intervention, their applications remain limited.

    Region-based ACMs address major limitations of edge-based models, such as noise sensitivity and weak boundary leakage, by leveraging region-based statistical information. The Mumford-Shah functional \cite{mumford1989optimal} established the theoretical foundation, but the model's high complexity poses significant challenges for numerical implementation. The seminal Chan-Vese (C-V) model \cite{chan2001active} enabled efficient two-phase segmentation of homogeneous regions. Subsequent variants \cite{zhang2010active,li2011level,ding2018active,niaz2023edge,FRAGL} addressed intensity inhomogeneity by employing either localized or global statistics.
	
	Deep convolutional networks have profoundly advanced image segmentation, beginning with U-Net's encoder-decoder skip architecture for spatial detail retention~\cite{ronneberger2015u}, extended by Deeplabv3+ via atrous pooling and multi-scale feature capture~\cite{chen2018encoder}, and further generalized through the promptable Segment Anything Model for cross-domain adaptability~\cite{zhang2024segment}. Zhao et al. enhanced this lineage by integrating star-shape constraints via convex-combination smoothing to improve stability~\cite{zhao2025convex}.

    Hybrid deep ACMs (Deep ACMs) effectively bridge data-driven and variational methods by embedding learnable features into contour evolution. Representative works include: Marcos et al.~\cite{marcos2018learning} with self-tuning ACMs, Zhang et al.~\cite{zhang2020deep} combining DenseUNet and C-V models, Zhao et al.~\cite{zhao2023attractive} introducing morphology-aware networks with contour attraction terms, and Jin et al.~\cite{jin2024regularized} developing regularized CNNs using geodesic active contours and iterative thresholding for noise robustness.
	
	Although CNNs excel in image segmentation, its heavy reliance on labeled data poses significant limitations. In contrast, traditional models like ACMs require no annotated data, offering greater efficiency and lower operational costs. The success of Deep ACMs further highlights the value of traditional models for developing interpretable segmentation frameworks. 

The hyperbolic mean curvature flow (HMCF), originally introduced by Yau \cite{yau2000review}, offers a mathematically grounded approach for controlling curve evolution. Kusumasari et al. \cite{kusumasari2018hyperbolic} later demonstrated its effectiveness in handling motion with obstacles. In this work, we systematically investigate the integration of HMCF into image segmentation within the partial differential equation (PDE) framework. The main contributions of this paper are threefold.
	\begin{itemize}
		\item We pioneer the incorporation of HMCF into image segmentation tasks. The proposed HMCF-ACMs enable adaptive selection of velocity fields and initial contours, establishing a unified PDE framework suitable for diverse image types without requiring redesign of the variational model. This offers greater flexibility compared to parabolic mean curvature flow-driven ACMs (PMCF-ACMs), as formalized in Eq.~\eqref{hmcf_apply}.
        \item We rigorously elucidate the differences between HMCF-ACMs and PMCF-ACMs (Eq.~\eqref{eq:v}), showing that the hyperbolic formulation naturally eliminates the zig-zag phenomenon inherent in gradient-descent schemes. This is achieved through an acceleration field that enables adaptive curvature-driven evolution. In comparison, the distance regularization term proposed by Li et al. \cite{li2010distance} only mitigates this issue and requires careful tuning of an additional hyperparameter.
        \item We demonstrate that HMCF-ACMs deliver superior denoising and lower parameter sensitivity relative to PMCF-ACMs under high-intensity noise. Extensive experiments on natural and medical images confirm this advantage (Section~\ref{sec:E}).
	\end{itemize}

	\section{Related Work}
	Let \( \Omega \subset \mathbb{R}^2 \) denote a bounded open domain, with boundary \( \partial \Omega \). Here, \(\vec{n}\) denotes the unit inward normal vector along \( \partial \Omega \). Let \( I: \overline{\Omega} \rightarrow \mathbb{R} \) represent a given image intensity function defined on the closure of \( \Omega \), and let \( C(s,t): \mathbb{S}^1 \times [0,T) \rightarrow \mathbb{R}^2 \) be a family of parametrized curves evolving over time \( t \) within this domain.
	
\begin{figure}[!t]
		\centering
		\includegraphics[width=2.5in]{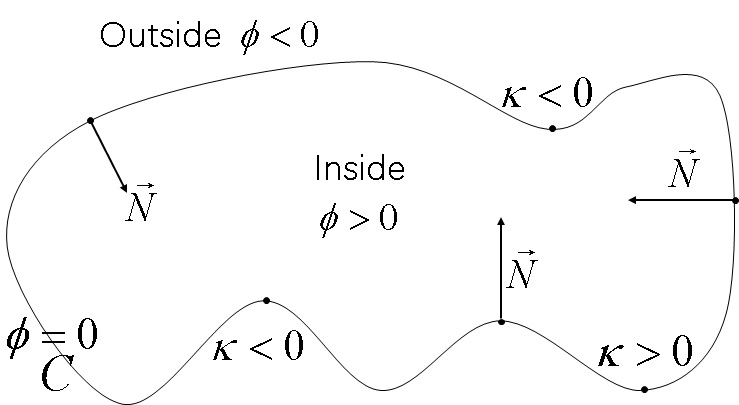}
		\caption{The curve \( C(t) = \{ (x, y) \in \Omega \mid \phi(x, y, t) = 0 \} \) propagates in the normal direction. For convex parts, the curvature \(\kappa \textgreater 0\), while for concave parts, the curvature \(\kappa \textless 0\).}
		\label{fig1}
	\end{figure}

	In curve evolution models based on the level set method, the curve \( C(t) \) is implicitly represented by the zero level set of the function \( \phi(x, y, t) \), defined as \( C(t) = \{ (x, y) \in \Omega \mid \phi(x, y, t) = 0 \} \). In this paper, we define the implicit function \( \phi(x, y) \) such that \( \phi(x, y)\textgreater0 \) in the interior region, \( \phi(x, y) \textless 0 \) in the exterior region, and \( \phi(x, y) = 0 \) on the curve. This spatial configuration is visually demonstrated in Fig. \ref{fig1}. To optimize computational efficiency and numerical stability, the level set function is commonly regularized as a  signed distance function (SDF). 
	
	As we know, PDE-based frameworks have primarily encompassed the curvature (or diffusive) term, the advection (or transport) term, and propagation (or expansion ) term for image segmentation \cite{osher2004level}. The general equation is
	\begin{equation}
		\frac{\partial \phi}{\partial t}(x,y,t) =\left(\kappa(x,y,t)+r(x,y)+u\right)|\nabla \phi|,
		\label{GeneralLevel}
	\end{equation}
	where:
\begin{itemize}
	\item \(\kappa |\nabla \phi|\) is the mean curvature term, where $\kappa$ is naturally extended as the curvature operator acting on all level sets. This term serves to smooth the curve, and reduce the occurrence of singular points.
	\item \(r|\nabla \phi|\) represents the advection term, where \(r\) is an externally generated velocity field that drives the contour to progressively converge toward the target interface.
	\item The propagation term  \(u|\nabla \phi|\) characterizes the evolution of a contour driven by a constant value. This term induces a constant motion in the normal direction, dictating either the contraction or expansion of the curve.
\end{itemize}

	The variational level set framework employs the Heaviside function \( H(\phi) \) to reformulate curve evolution as energy functional minimization. For numerical stability, regularized approximations of \( H(\phi) \) and its distributional derivative (the Dirac delta function \( \delta(\phi) \)) are usually given by
	\begin{equation}
			H_\epsilon(\phi) = \frac{1}{2}\left(1 + \frac{2}{\pi}\arctan\left(\frac{\phi}{\epsilon}\right)\right), \;
		\delta_\epsilon(\phi) = \frac{1}{\pi}\frac{\epsilon}{\epsilon^2 + \phi^2},
	\end{equation}

	where the regularization parameter \(\epsilon \textgreater 0\) controls the interface transition width. Now we are going to briefly review some models.

	\subsection{GAC Model}
	Caselles et al. \cite{caselles1997geodesic} proposed the geodesic GAC model, an intrinsic formulation that is independent of the curve's parameterization.
	
	The proposed energy functional of GAC model is
	\begin{equation}
		\int_{\mathbb{S}^1} g(|\nabla I(C(s))|) |C'(s)| \, ds.
		\label{GAC_C}
	\end{equation}
	The corresponding gradient descent flow is given by
	\begin{equation}
		\frac{\partial C}{\partial t} = \left( \kappa g - \langle \nabla g, \vec{N} \rangle \right) \vec{N},
		\label{GAC_F}
	\end{equation}
	where \(\vec{N}\) denoting the inward unit normal vector.
	By embedding the level set function into Eq. \eqref{GAC_F}, we obtain
	\begin{equation}
		\frac{\partial \phi}{\partial t} = g\,  \text{div}  \left(\frac{\nabla \phi}{|\nabla \phi|} \right)  |\nabla \phi| + \langle \nabla g, \nabla \phi \rangle.
	\end{equation}
	
	By employing the variational level set method, the energy functional \eqref{GAC_C} can be reformulated as
	\begin{equation}
	E_{GAC}(\phi) = \int_{\Omega} g(|\nabla I|) \delta_{\epsilon}(\phi) |\nabla \phi| \, dxdy.
\end{equation}
	Then,  the gradient descent flow is expressed as 
	\begin{equation}
	\frac{\partial \phi}{\partial t} = g\,\text{div} \left( \frac{\nabla \phi}{|\nabla \phi|} \right)\delta_{\epsilon}(\phi) + \langle \nabla g, \nabla \phi \rangle.
\end{equation}
	
	To address the challenge of segmenting images with deep concave regions, an enhanced GAC model introduces a propagation term to ensure the curve continuous evolving. The evolution equation is given by
	\begin{equation}
		\frac{\partial \phi}{\partial t} = g\,\delta_{\epsilon}(\phi)\left[\text{div} \left( \frac{\nabla \phi}{|\nabla \phi|} \right)+u\right] + \langle \nabla g, \nabla \phi \rangle.
		\label{GAC_V}
	\end{equation}
	Here, \( u \geq 0 \) is a constant, and it satisfies \(\kappa + u \textgreater 0.\)
	\subsection{C-V Model}
	For images with approximately constant intensity, the segmentation task is formulated as identifying an optimal closed contour \( C \) that partitions the image domain \( \Omega \) into interior (\( \omega \)) and exterior (\( \Omega \setminus \omega \)) regions with statistically homogeneous intensity variances. This foundational framework was established by Chan and Vese~\cite{chan2001active} through their region-based ACM. To eliminate the need for reinitialization, the distance regularization term proposed by Li et al.~\cite{li2010distance} is integrated into the energy functional. The augmented energy functional is defined as
\begin{equation}
\begin{aligned}
		E_{CV}(\phi, c_1, c_2) &= \mu \int_{\Omega} \delta_{\epsilon}(\phi) |\nabla \phi| \, dxdy\\
		&+ \lambda \int_{\Omega} (I - c_1)^2 H_{\epsilon}(\phi) \, dxdy \\
		&+ \lambda \int_{\Omega} (I - c_2)^2 (1 - H_{\epsilon}(\phi)) \, dxdy \\
		&+\frac{\gamma}{2} \int_{\Omega} \left( |\nabla \phi| - 1 \right)^2 \, dx \, dy.
	\end{aligned}
\end{equation}

	Using the steepest descent method, the energy functional \( E_{CV}\) with respect to \( c_1 \) and \( c_2 \) minimizes to
	\begin{equation}
\begin{aligned}
	c_1 = \frac{\int_{\Omega} I(x,y) H_{\epsilon}(\phi) \, dxdy}{\int_{\Omega} H_{\epsilon}(\phi) \, dxdy},
	c_2 = \frac{\int_{\Omega} I(x,y) [1 - H_{\epsilon}(\phi)] \, dxdy}{\int_{\Omega} [1 - H_{\epsilon}(\phi)] \, dxdy}.
\end{aligned}
	\end{equation}
	Fixing \( c_1 \) and \( c_2 \), the energy functional \( E_{CV} \) is minimized with respect to \( \phi \), yielding the evolution equation
	\begin{equation}
		\begin{aligned}
			\frac{\partial \phi}{\partial t} &= \delta_{\epsilon}(\phi) \left[ \mu\, \text{div} \left( \frac{\nabla \phi}{|\nabla \phi|} \right) - \lambda (I - c_1)^2 + \lambda (I - c_2)^2 \right] \\
			&\quad + \gamma \left[ \nabla^2 \phi -\text{div} \left( \frac{\nabla \phi}{|\nabla \phi|}\right)\right],
		\end{aligned}
		\label{CV_V}
	\end{equation}
	where \(\mu\textgreater0\) and \(\gamma \textgreater 0\) are constants, and \(\nabla^2\) is the Laplacian operator.

	\section{Proposed Methodology}
	\subsection{Lagrangian Formulation of HMCF Model}
	As we know, the PMCF is inherently constrained by having only one degree of freedom in its formulation --- specifically, its solution is uniquely determined by the following initial value problem
	\begin{equation}
		\left\{
		\begin{aligned}
			&\frac{\partial C}{\partial t} = \mu\kappa \vec{N}, \quad \text{in } \mathbb{S}^1 \times [0, T),\\
			&C(s, 0) = C_0,\\
		\end{aligned}
		\right.
		\label{eq:pmcf}
	\end{equation}
	where
	\begin{itemize}
		\item \( \mu \textgreater 0 \) modulates curvature effects;
		\item \(C_0\) represents the initial position of curve.
	\end{itemize}
	This parabolic normal flow can be discretized and solved iteratively using gradient descent
	\begin{equation}
		C(s, \tau_k) = C(s,\tau_{k-1}) + \tau \mu \kappa(s,\tau_{k-1})\vec{N}(s,\tau_{k-1}), \label{eq:discrete_C} 
	\end{equation}
	with temporal partitioning defined as
	\[
	\begin{aligned}
		\tau &:= T/M, \quad M \in \mathbb{Z}^+,  \\
		\tau_k &:= k\tau, \quad k \in \{0,1,\ldots,M\}. 
	\end{aligned}
	\label{eq:tau_def}
	\]
	
	Eq. \eqref{eq:discrete_C} implies that within each interval \([\tau_{k-1}, \tau_k)\), the curve \( C^k\) evolves from \(C(s,\tau_{k-1})\) at a fixed velocity \(\mu\kappa(s,\tau_{k-1})\) along \(\vec{N}(s,\tau_{k-1})\), that is,
	\begin{equation}
		\frac{\partial C^k}{\partial t} = \mu\kappa(s,\tau_{k-1})\vec{N}(s,\tau_{k-1}).
		\label{eq:fixed_velocity}
	\end{equation}
	
	Yet this method suffers from zig-zag instability \cite{dong2021class}: when a curve’s curvature varies sharply, a fixed step (or hand-tuned~\(\mu\)) cannot adapt simultaneously to all points—it overshoots where curvature is high and stalls where it is low, producing oscillations.
	
	To overcome this gradient-descent drawback, we introduce a certain HMCF --- a curve evolution scheme  driven by a self-generated acceleration field
	\begin{equation}
		\vec{F}(s,t) := b\kappa\vec{N} - \left\langle \frac{\partial^2 C}{\partial s \partial t}, \frac{\partial C}{\partial t} \right\rangle \vec{T}.
		\label{eq:acceleration_field}
	\end{equation}
	The corresponding initial value problem for this HMCF model is formulated as
	\begin{equation}
		\left\{
		\begin{aligned}
			&\frac{\partial^2 C}{\partial t^2}(s, t) = \vec{F},\\
			&C(s, 0) = C_0, \\
			&\frac{\partial C}{\partial t}(s, 0) = v_0\vec{N}_0,
		\end{aligned}
		\right.
		\label{eq:hmcf}
	\end{equation}
	where
	\begin{itemize}
		\item \( b \textgreater 0 \) modulates curvature effects,
		\item  \( \vec{T} \) denotes the unit tangent vectors of \(C\),
		\item \( v_0 \vec{N}_0 \) defines the initial velocity field.
	\end{itemize}
	
	The local existence and uniqueness of solutions for \eqref{eq:hmcf} have been rigorously established by Kong et al. \cite{dexing2009hyperbolic}.
	
	\theoremheader{Theorem 1 \cite{dexing2009hyperbolic}} 
	Let \(C_0\) be a smooth strictly convex closed curve. Then, there exists a positive constant \( \hat{T} \) and a family of strictly convex closed curves \( \{C(\cdot, t)\}_{t \in [0, \hat{T})} \) satisfying \eqref{eq:hmcf}, provided that the initial velocity \( v_0(s) \) is smooth on \( \mathbb{S}^1 \).
	
	\theoremheader{Theorem 2 \cite{dexing2009hyperbolic}} 
	Let \(C_0\) be a smooth strictly convex closed curve. Then, there exists a class of initial velocities such that the solution to \eqref{eq:hmcf}  exists only over a finite time interval \([0, \hat{T}_{\max})\). Furthermore, the solution \( C(\cdot, t) \) converges to a single point as \( t \to \hat{T}_{\max} \).
	
	These theorems provide rigorous mathematical guarantees for the feasibility of our HMCF framework in image segmentation tasks.
	
	An additional characteristic of the HMCF framework lies in its inherent geometric consistency. Specifically, when the initial velocity field is oriented along the normal direction, the evolving curve \( C(s,t) \) exhibits strictly normal motion throughout the entire evolution process. Mathematically, this is expressed as
	\begin{equation}
		\left\langle \frac{\partial C}{\partial t}, \frac{\partial C}{\partial s} \right\rangle = 0, \quad \forall (s,t) \in \mathbb{S}^1 \times [0,\hat{T}_{\max}).
	\end{equation}
	To demonstrate this property, we consider that the curve \(C\) satisfies Eq. \eqref{eq:hmcf}. Hence, we have
	\begin{equation}
		\begin{aligned}
		&\frac{\partial}{\partial t}\left\langle \frac{\partial C}{\partial t},\frac{\partial C}{\partial s}\right\rangle=\left\langle \frac{\partial^2 C}{\partial t^2},\frac{\partial C}{\partial s}\right\rangle+\left\langle \frac{\partial C}{\partial t},\frac{\partial^2 C}{\partial t\partial s}\right\rangle \\
		&=\left\langle b\kappa\vec{N}-\left\langle \frac{\partial^2 C}{\partial s \partial t}, \frac{\partial C}{\partial t} \right\rangle \vec{T},\vec{T} \right\rangle+\left\langle \frac{\partial C}{\partial t},\frac{\partial^2 C}{\partial t\partial s}\right\rangle \\
		&=-\left\langle \frac{\partial^2 C}{\partial s \partial t}, \frac{\partial C}{\partial t} \right\rangle+\left\langle \frac{\partial C}{\partial t},\frac{\partial^2 C}{\partial t\partial s}\right\rangle =0.
	\end{aligned}
	\end{equation}
	Thus, for any $ t\in [0,\hat{T}_{\max})$, it holds that
	\begin{equation}
		\left\langle \frac{\partial C}{\partial t}, \frac{\partial C}{\partial s} \right\rangle = \left\langle \frac{\partial C}{\partial t}, \frac{\partial C}{\partial s} \right\rangle \bigg|_{t=0} = \left\langle v_0 \vec{N}_0, \vec{T}_0 \right\rangle = 0,
	\end{equation}
	where \(\vec{T}_0\) is the initial unit tangent vector. Consequently, the velocity field maintains strict normal orientation, i.e.,
	\begin{equation}
		\frac{\partial C}{\partial t} = v(s,t)\vec{N}(s,t).
		\label{eq:normal_velocity}
	\end{equation}
	
	To rigorously derive the analytical expression for the magnitude of the velocity field \(v(s,t)\), we differentiate Eq. \eqref{eq:normal_velocity} with respect to variables $t$ and $s$, thereby obtaining the following equations
	\begin{equation}
		\left\{
	\begin{aligned}
		&\frac{\partial^2 C}{\partial t^2} = \frac{\partial v}{\partial t}\vec{N}+v\cdot \frac{\partial \vec{N}}{\partial t},\\
		&\frac{\partial^2 C}{\partial t \partial s} = \frac{\partial v}{\partial s}\vec{N}+v\cdot \frac{\partial \vec{N}}{\partial s}.
	\end{aligned}
	\right.
	\end{equation}
	Using the formulas \(\left\langle \vec{N},\vec{T}\right\rangle=0\) and \(\vec{T}=\frac{\partial C}{\partial s}\), we obtain
	\begin{equation}
		\begin{aligned}
		\left\langle \frac{\partial \vec{N}}{\partial t},\vec{T}\right\rangle&=-\left\langle \vec{N},\frac{\partial \vec{T}}{\partial t} \right\rangle=-\left\langle \vec{N},\frac{\partial^2 C}{\partial t\partial s} \right\rangle \\
		&=-\left\langle \vec{N}, \frac{\partial v}{\partial s}\vec{N}+v\cdot \frac{\partial \vec{N}}{\partial s}\right\rangle \\
		&=-\frac{\partial v}{\partial s}-\left\langle \vec{N}, v\cdot \frac{\partial \vec{N}}{\partial s}\right\rangle.
	\end{aligned}
	\end{equation}
	Then, by applying the Frenet formulas
	\begin{equation}
		\frac{\partial \vec{T}}{\partial s} = \kappa \vec{N}, \quad \frac{\partial \vec{N}}{\partial s} = -\kappa \vec{T},
	\end{equation}
	we deduce that
	\begin{equation}
		\left\langle \frac{\partial \vec{N}}{\partial t},\vec{T}\right\rangle=-\frac{\partial v}{\partial s}+\left\langle \vec{N},v\kappa\vec{T}\right\rangle=-\frac{\partial v}{\partial s}.
	\end{equation}
	Thus, the acceleration fields simplify to
	\begin{equation}
		\begin{aligned}	
		\frac{\partial^2 C}{\partial t^2} &=\frac{\partial v}{\partial t}\vec{N}+v \left\langle \frac{\partial \vec{N}}{\partial t},\vec{N} \right\rangle\vec{N}+v \left\langle \frac{\partial \vec{N}}{\partial t},\vec{T} \right\rangle\vec{T} \\
		&=\frac{\partial v}{\partial t}\vec{N}-v\cdot\frac{\partial v}{\partial s}\vec{T}.
	\end{aligned}
	\end{equation}
By utilizing  \eqref{eq:hmcf}, we  establish the following important  identities 
	\begin{equation}
		\frac{\partial v}{\partial t}=b\kappa, \quad
	v\frac{\partial v}{\partial s}=\left\langle \frac{\partial^2 C}{\partial s \partial t}, \frac{\partial C}{\partial t} \right\rangle.
	\end{equation}
	Under the initial condition \(\frac{\partial C}{\partial t}(s, 0) = v_0\vec{N}_0,\) we derive that the magnitude of the velocity field should evolve as
	\begin{equation}
		v(s,t) = v_0 + b\int_0^t \kappa(s,\xi)d\xi.
		\label{eq:v}
	\end{equation}
	
	In summary, the HMCF framework (Eq.~\eqref{eq:hmcf}) constrains the velocity field to the normal direction (Eq.~\eqref{eq:normal_velocity}), consistent with the behavior of PMCF (Eq.~\eqref{eq:pmcf}). Eq.~\eqref{eq:v} reveals how the curvature-acceleration field directly sets the velocity magnitude, a mechanism fundamentally different from PMCF.
	
	Despite these advantages, the Lagrangian formulation introduces significant computational challenges \cite{osher2004level}. Topological changes necessitate dynamic discretization adjustments, and interface regularization requires carefully balanced dissipation. Current implementations face exponential complexity growth due to detachment-reconnection operations during topological transitions.

	\subsection{Eulerian Formulation of HMCF Model}
	
	To overcome  Lagrangian limitations, we construct an Eulerian framework defining the dynamic interface through time-dependent level set function
		\begin{equation}
			C(s,t) = \{(x,y) \mid \phi(x, y,t) = 0\}, \quad t \in [0, \hat{T}_{\max}).
		\end{equation}
	The evolution of \(\phi\) governs the interface motion and is described by
	\begin{equation}
		\frac{d\phi}{dt} = \frac{\partial \phi}{\partial t} + \nabla \phi \cdot \vec{V} = 0,
		\label{eq:transport}
	\end{equation}
	which is derived from the principle of total differentiation, with \(\vec{V}\) denotes the velocity field. From the first equation in \eqref{eq:hmcf}, we have \(\vec{F}	=\frac{d \vec{V}}{dt}\). Then, the temporal differentiation of Eq. \eqref{eq:transport} leads to
	\begin{equation}
		\frac{\partial^2 \phi}{\partial t^2}+\nabla(\frac{\partial \phi}{\partial t})\cdot\vec{V}=-\frac{d}{dt}(\nabla \phi)\cdot\vec{V}-\vec{F}\cdot\nabla \phi.
		\label{hmcf_EE1}
	\end{equation}
	From the preservation property of normal motion  in \eqref{eq:normal_velocity}, the velocity field satisfies
	\begin{equation}
		\frac{\partial C}{\partial t} = \vec{V} = v\vec{N}, \quad t \in [0, \hat{T}_{\max}),
		\label{ortho_velocity}
	\end{equation}
	where $\vec{N} = \nabla\phi/|\nabla\phi|$. Substituting this geometric identity into \eqref{hmcf_EE1}, we obtain
	\begin{equation}
		\begin{aligned}
			\left\langle \nabla\left(\frac{\partial \phi}{\partial t}\right), \vec{V} \right\rangle &= \frac{v}{|\nabla\phi|} \left\langle \frac{\partial}{\partial t}(\nabla\phi), \nabla \phi \right\rangle, \\
			\left\langle \frac{d}{dt}(\nabla \phi), \vec{V} \right\rangle &= \frac{v}{|\nabla\phi|} \left\langle \frac{d}{dt}(\nabla\phi), \nabla \phi \right\rangle.
		\end{aligned}
	\end{equation}
	
	When the level set function $\phi$ is maintained as a SDF, substantial computational simplifications emerge due to its inherent geometric properties. The SDF $d(x,y,t)$, formally defined as the minimal signed Euclidean distance from any spatial point $(x,y)$ to the evolving interface $\Gamma_t$, is expressed as
		\begin{equation}
			d(x,y,t) = 
		\begin{cases} 
			\underset{(m,n) \in \Gamma_t }{\inf}\|(x,y) - (m,n)\|,
			&\text{if } (x,y) \in \omega_t, \\
			\underset{(m,n) \in \Gamma_t }{-\inf}\|(x,y) - (m,n)\|,
			&\text{otherwise},
		\end{cases}
		\end{equation}
	where $\omega_t = \{(x,y) \in \Omega \mid \phi(x,y,t) \textgreater 0\}$ denotes the interior region bounded by $\Gamma_t = \partial \omega_t$. 
	
	The gradient normalization condition \( |\nabla d| \equiv 1 \) holds for all \( t \in [0, \hat{T}_{\max}) \) at points where the closest projection onto \(\Gamma_t\) is unique. However, this property does not hold for points that are equidistant from at least two distinct points on the interface \(\Gamma_t\) \cite{osher2004level}. Fortunately, this limitation does not affect the effectiveness in our numerical computations.
	
	Leveraging this normalization constraint $(\nabla \phi \cdot \nabla \phi = 1)$, differentiation with respect to time yields two critical orthogonality relations
	\begin{equation}
		\left\langle \frac{\partial}{\partial t}(\nabla \phi), \nabla \phi \right\rangle = 0 \quad \text{and} \quad \left\langle \frac{d}{dt}(\nabla \phi), \nabla \phi \right\rangle = 0.
		\label{eq:orthogonality}
	\end{equation}
	Substituting \eqref{eq:orthogonality} into \eqref{hmcf_EE1} eliminates tangential components, leading to the simplified hyperbolic equation
	\begin{equation}
		\frac{\partial^2 \phi}{\partial t^2} = -\vec{F} \cdot \nabla \phi.
		\label{level_F}
	\end{equation}
	
	Since \(\vec{N}\) and \(\nabla\phi\) have the same direction, we have \(\vec{T} \cdot \nabla\phi = 0\), which implies that the tangential acceleration components vanish when Eq. \eqref{eq:acceleration_field} is plugged into Eq. \eqref{level_F}. Thus, Eq. \eqref{level_F} can be rewritten as
	\begin{equation}
		\frac{\partial^2 \phi}{\partial t^2} = -b \kappa.
		\label{level_F_rewritten}
	\end{equation}
	Furthermore, noting that the level set formulation for curvature is given by 
	\begin{equation}
		\kappa = -\text{div} \left( \frac{\nabla \phi}{|\nabla \phi|} \right),
		\label{curvature}
	\end{equation}
	we can obtain the following wave equation
	\begin{equation}
		\frac{\partial^2 \phi}{\partial t^2} = b \nabla^2 \phi.
		\label{hmcf_FB}
	\end{equation}
	As a consequence, the complete initial-boundary value problem becomes
	\begin{equation}
		\left\{
		\begin{aligned}
			&\frac{\partial^2 \phi}{\partial t^2} = b \nabla^2 \phi, \hspace{6.3em}\text{in } \Omega \times [0, \hat{T}_{\max}), \\
			&\frac{\partial \phi}{\partial \vec{n}} = 0, \hspace{8.5em}\text{on } \partial \Omega \times [0, \hat{T}_{\max}), \\
			&\frac{\partial \phi}{\partial t}(x,y,0) =-v_0, \hspace{4.2em}\text{in } \Omega, \\
			&\phi(x, y,0) = d(x,y,0), \hspace{3em}\text{in } \Omega.
		\end{aligned}
		\right.
		\label{eq:hmcf_E}
	\end{equation}
	
	In conclusion, Eqs. (\ref{eq:hmcf}) and (\ref{eq:hmcf_E}) produce identical outcomes in terms of curve positioning, provided that the function $\phi$ is maintained as the SDF. 
	
	\subsection{Energy Functional}
	The first equation in Eq. \eqref{eq:hmcf_E} can be interpreted as a vibrating membrane system \cite{kusumasari2018hyperbolic}, where the total mechanical energy combines surface tension and kinetic components, i.e., 
		\begin{equation}
			\int_{0}^{\hat{T}_{\max}} \int_{\Omega} \left[\left (\frac{\partial \phi}{\partial t}\right)^2 - b|\nabla \phi|^2 \right] dxdy dt.
		\end{equation}
	Minimization of this action functional yields the hyperbolic evolution equation
	\begin{equation}
		\frac{\partial^2 \phi}{\partial t^2} = b\nabla^2 \phi,\quad \text{in } \Omega \times [0,\hat{T}_{\max}),
		\label{wave_equation}
	\end{equation}
	which is of course consistent with \eqref{eq:hmcf_E}.
	\section{IMPLEMENTATION}
	\subsection{HMCF-ACM}	
	As governed by Eq. \eqref{eq:fixed_velocity}, which prescribes fixed-velocity contour evolution over discrete intervals \([\tau_{k-1},\tau_k)\), we introduce the HMCF-structured acceleration field \eqref{eq:acceleration_field} to these temporal partitions. Subsequently, the Eulerian formulation described by Eq. \eqref{eq:hmcf_E} is implemented for temporal evolution across each interval \([\tau_{k-1},\tau_k)\). Within this framework, the evolving contour  \(C^k(s,t)\) is implicitly represented as
	\begin{equation}
		C^k(s,t) = \{(x,y) \mid \phi^k(x,y,t) = 0\}, \quad t \in [0, \tau),
	\end{equation}
	where \(\phi^k\) evolves according to the hyperbolic initial-boundary value problem
	\begin{equation}
		\left\{
		\begin{aligned}
			&\frac{\partial^2 \phi^k}{\partial t^2} = b\nabla^2\phi^k,\hspace{5em}  \text{in } \Omega\times [0, \tau), \\
			&\frac{\partial \phi^k}{\partial \vec{n}} = 0, \hspace{7.7em} \text{on } \partial I \times [0, \tau), \\
			&\frac{\partial \phi^k}{\partial t}(x,y,0) = v_0^k(x,y),\hspace{1.6em} \text{in } \Omega, \\
			&\phi^k(x,y,0) = d^{k-1}(x,y), \hspace{1.5em} \text{in } \Omega,
		\end{aligned}
		\right.
		\label{hmcf_apply}
	\end{equation}
	where
	\begin{itemize}
		\item \textbf{Configurable Initial Velocity} \(v_0^k\): a task-specific value  manually selected to achieve different segmentation objectives.
		
		\item \textbf{SDF} \(d^{k-1}\): a projection operator based on \(\phi^{k-1}(x,y,\tau)\), indeed
		\begin{equation}
			d^{k-1}(x,y) = 
			\begin{cases} 
			\underset{(m,n) \in \Gamma^{k-1}}{\inf} \|(x,y) - (m,n)\|,\\
				\text{if } (x,y) \in \omega^{k-1} \\
				\underset{(m,n) \in \Gamma^{k-1}}{-\inf} \|(x,y) - (m,n)\|,\\
				\text{otherwise},
			\end{cases}
		\label{re}
		\end{equation}
		in which \(\omega^{k-1} = \{(x,y) \in \Omega \mid \phi^{k-1}(x,y,\tau) \textgreater 0\}\) and \(\Gamma^{k-1} = \partial\omega^{k-1}\).
	\end{itemize}
	By coupling velocity magnitude to the curvature-acceleration field, HMCF injects second-order dynamics that eliminate the drawback of gradient descent scheme \eqref{eq:discrete_C}.
	\begin{remark}[SDF Maintenance]
		\label{rmk:sdf_maintenance}
		During temporal evolution, \(\phi^k\) will generally drift away from its initialized value as signed distance. We employ
		\begin{itemize}
			\item \textbf{Periodic reinitialization}: project \(\phi^k\) to the SDF space using Eq. \eqref{re},
			\item \textbf{Distance regularization}: augment the velocity with penalty term \( \gamma \left[ \nabla^2 \phi -\text{div} \left( \frac{\nabla \phi^k}{|\nabla \phi^k|}\right) \right]\) to preserve numerical stability.
		\end{itemize}
	\end{remark}
	
\begin{figure}[!t]
		\centering
		\includegraphics[width=3.5in]{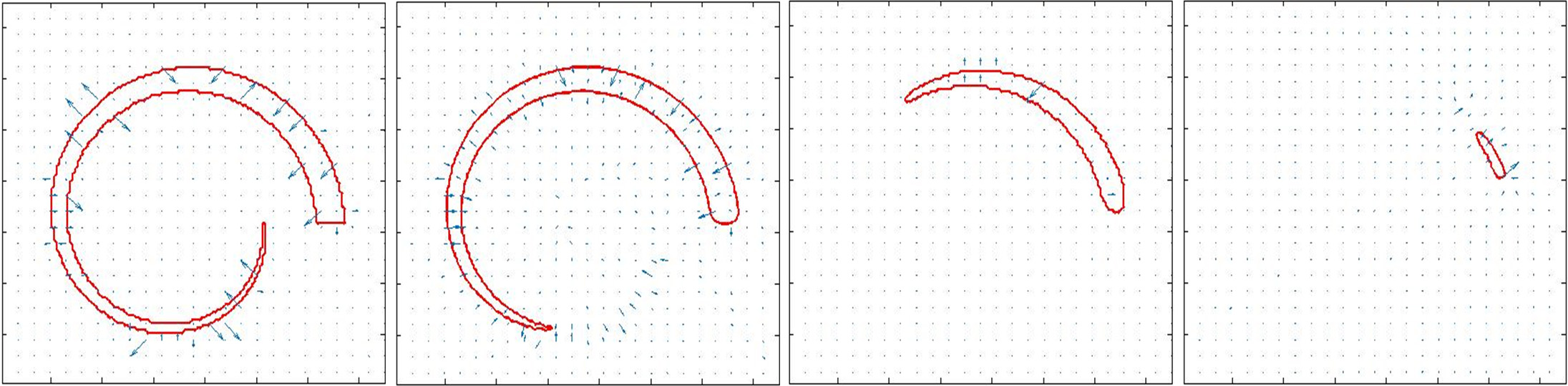}
		\caption{Curvature-driven spiral evolution under the HMCF. Left to right: initial contours (red circles), intermediate evolution stages, and final result. Acceleration fields \( \vec{F}^k(x,y,0) \) at iterations \( k = \{0, 2, 10, 22\} \).}
		\label{wound}
	\end{figure}

\begin{figure}[!t]
		\centering
		\includegraphics[width=3.5in]{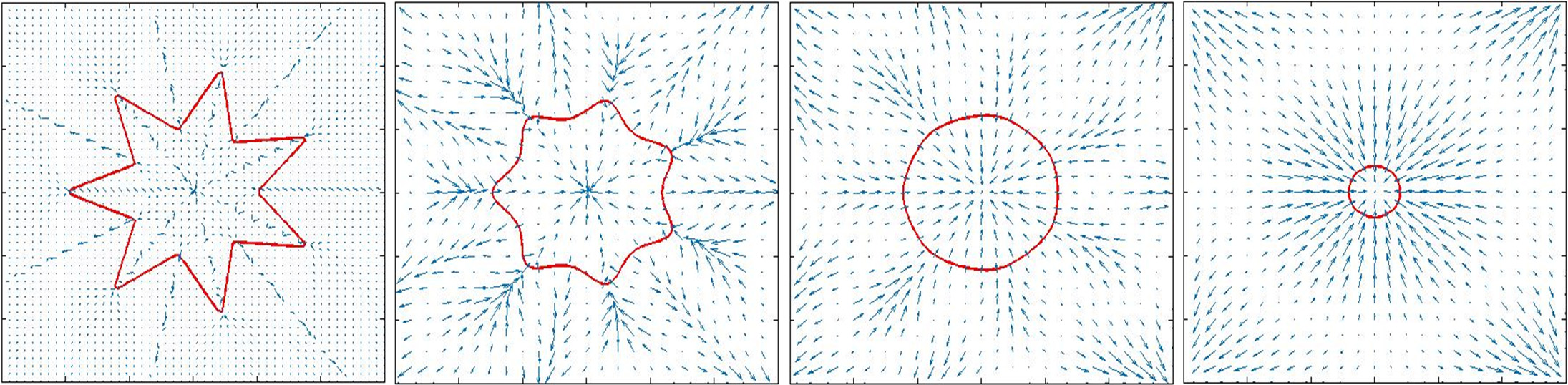}
		\caption{Curvature-driven star evolution under the HMCF. Acceleration fields \( \vec{F}^k(x,y,0) \) at iterations \( k = \{0, 20, 75, 100\} \).}
		\label{star}
	\end{figure}

	To visually elucidate the geometric regularization effect of \eqref{hmcf_apply}, we present two canonical cases with null initial velocities ($v_0^k = 0$):
\begin{itemize}
		\item Spiral Evolution (Fig. \ref{wound}): high-curvature terminal regions of the wound spiral exhibit accelerated motion under $\vec{F}^k$, contrasting with the quasi-static low-curvature section.
		
		\item Star Evolution (Fig. \ref{star}): curvature-driven flow induces tip inversion and inter-tip expansion, demonstrating bidirectional topology changes.
	\end{itemize}
	
	The subsequent discussion focuses on the specific configuration of velocity-fields for HMCF-ACMs. Notably, these configurations exclude the curvature term while preserving the data fidelity term.
	\subsubsection{HMCF-Driven Chan-Vese Model~\cite{chan2001active}}
	The HMCF-CV model ensures global convergence on piecewise-constant images by employing the velocity field
	\begin{equation}
		v_0^k = \delta_\epsilon(\phi^{k}(0))\left[(I - c_2^k)^2 - (I - c_1^k)^2\right].
	\end{equation}
	
	\subsubsection{HMCF-Driven Local Binary Fitting Model~\cite{li2011level}}
    The HMCF-LBF model handles intensity inhomogeneity via a bias field $J(\mathbf{x})$. The initial velocity reads
    \begin{equation}
    v_0^k=\delta_\varepsilon\!\bigl(\phi^k(0)\bigr)\bigl(e_2^k-e_1^k\bigr),
    \end{equation}
    with
    \begin{equation}
    e_i^k(\mathbf{x})=\int_\Omega K_\sigma(\mathbf{y}-\mathbf{x})\bigl|I(\mathbf{y})-J(\mathbf{x})c_i^k\bigr|^2d\mathbf{y},\quad i=1,2,
    \end{equation}
    where the Gaussian kernel
    \begin{equation}
    K_\sigma(\mathbf{y}-\mathbf{x})=\frac{1}{2\pi\sigma^2}\exp\!\Bigl(-\frac{\|\mathbf{y}-\mathbf{x}\|^2}{2\sigma^2}\Bigr)
    \end{equation}
    controls the locality scale.
    
    \subsubsection{HMCF-Driven Local Pre-Fitting Model~\cite{ding2018active}}
    The HMCF-LPF model addresses intensity variations via a pre-computed fitting function  $f_s$ and $f_l$,
    \begin{equation}
    v_0^k=\delta_\varepsilon\!\bigl(\phi^k(0)\bigr)\bigl(e_l-e_s\bigr),
    \end{equation}
    where
    \begin{equation}
		\begin{cases}
		e_s(\mathbf{x}) = \int_{\Omega} K_{\sigma}(\mathbf{y} - \mathbf{x}) |I(\mathbf{x}) - f_s(\mathbf{y})|^2 d\mathbf{y}, \\
		e_l(\mathbf{x}) = \int_{\Omega} K_{\sigma}(\mathbf{y} - \mathbf{x}) |I(\mathbf{x}) - f_l(\mathbf{y})|^2 d\mathbf{y}.
	\end{cases}
	\end{equation}
    
    \subsubsection{HMCF-Driven FRAGL Model~\cite{FRAGL}}
    HMCF-FRAGL fuses local and global statistics through the velocity
    \begin{equation}
		\begin{aligned}
			v_0^k &= \delta_{\varepsilon}(\phi^{k}(0)+\frac{1}{2})\left[( \phi^{k}(0)+\frac{1}{2})\left|I-(\alpha f_{o}+\beta c_{1}^{k})\right|^{2}\right. \\
			&\quad \left. -(\frac{1}{2}-\phi^{k}(0))\left|I-(\alpha f_{b}+\beta c_{2}^{k})\right|^{2}\right],
		\end{aligned}
	\end{equation}
    with $f_o,f_b$ the local object/background means and $\alpha+\beta=1$ balancing local and global fitting energies.
	
	Since the PMCF term is discarded, its spatial-discretisation stability constraint is lifted and the iteration time step $\tau$ can be chosen more freely.
		
	\subsection{Numerical Scheme}
	The numerical solution of the hyperbolic system \eqref{hmcf_apply} is implemented through a weighted fourth-order Runge-Kutta (R-K) temporal integration scheme \cite{ma2011nearly}. To further elaborate on the temporal discretization, the time step is selected as \(\hat{\tau} = \frac{\tau}{L}\), where \(L\) is a positive integer and \(0 \textless \hat{\tau} \leq \tau\). The \(l\)-th time step is denoted by \(\hat{\tau}_{l} = l\hat{\tau}\). The algorithm is presented as follows
	
	\begin{enumerate}
		\item  Transform  Eq. \eqref{hmcf_apply} into a system of first-order PDEs
		\begin{equation}
			\left\{
			\begin{array}{l}
				\frac{\partial Z^k}{\partial t} = D\cdot \phi^k, \\[2mm]
				\frac{\partial \phi^k}{\partial t} = Z^k.
			\end{array}
			\right.
			\label{rk1}
		\end{equation}
		where the differential operator \(D\) is defined as
		\begin{equation}
			D = b \left( \frac{\partial^2}{\partial x^2} + \frac{\partial^2}{\partial y^2} \right).
		\end{equation}
		\item  Rewrite the equations into a unified form.
		Introducing the notation \( \mathbf{W^k} = \left[ Z^k,\phi^k \right]^{\top} \),  Eq. \eqref{rk1} can be further rewritten as
		\begin{equation}
			\frac{\partial \mathbf{W^k}}{\partial t} = \mathbf{L} \cdot \mathbf{W^k},
			\label{rk2}
		\end{equation}
		where the differential operator \( \mathbf{L} \) is defined as
		\begin{equation}
			\mathbf{L} = \left[
		\begin{array}{cc}
			0 & D \\
			1 & 0
		\end{array}
		\right].
		\end{equation}
		\item The right-hand side of Eq. \eqref{rk2} is discretized into a semi-discrete system of ordinary differential equations. To compute the second-order spatial derivatives of the function $ \phi^k(x, y, t) $, we define $ \phi_{i,j}^{k,l} $ and $ Z_{i,j}^{k,l} $ as the function value and velocity at the grid point $ (x_i, y_j, \hat{\tau}_{l}) $, respectively. Here, $ x_i = i\Delta x $ and $ y_j = j\Delta y $, where $ \Delta x $ and $ \Delta y $ are the spatial step sizes. Without loss of generality, we assume $ \Delta x = \Delta y = 1 $. Unlike the approach in \cite{ma2011nearly}, which employs local interpolation-based approximations to handle second-order and three-order spatial derivatives separately, our method utilizes the Discrete Cosine Transform (DCT) for efficient frequency-domain derivative computation. Specifically, the derivatives are obtained by applying a two-dimensional DCT to the function values and multiplying by the corresponding wave numbers
		\begin{equation}
			\begin{aligned}
				D\mathbf{W}^{k}_{i,j} = \mathcal{DCT}^{-1} \left\{ -b(k_x^2 + k_y^2) \cdot \mathcal{DCT} \left\{ \mathbf{W}^{k}_{i,j} \right\} \right\},
			\end{aligned}
			\label{rk3}    
		\end{equation}
		where $ \mathcal{DCT} $ denotes the discrete cosine transform operator, $ k_x $ and $ k_y $ represent the wave numbers in the $ x $ and $ y $ directions, respectively, and $ \mathcal{DCT}^{-1} $ is the inverse transform. This approach capitalizes on the spectral accuracy of the DCT and is particularly well-suited for equations with Neumann boundary conditions.
		
		\item Discrete the time derivative by using a four-stage and four-order R-K. The calculation formula is given by
		\begin{equation}
			\left\{
			\begin{array}{ll}
				Z^{k,*}_{i,j}= Z^{k,l}_{i,j} + \frac{1}{2} \hat{\tau} \mathbf{D} Z^{k,l}_{i,j} + \frac{1}{4} \hat{\tau}^2 \mathbf{D} Z^{k,l}_{i,j},\\[2mm]
				\phi^{k,*}_{i,j}= \phi^{k,l}_{i,j} + \frac{1}{2} \hat{\tau}(\eta Z^{k,l}_{i,j}+(1-\eta)Z^{k,*}_{i,j})\\[2mm]
				\hspace{3em} + \frac{1}{4} \hat{\tau}^2 \mathbf{D} \phi^{k,l}_{i,j},\\[2mm]		
				\mathbf{W}^{k,l+1}_{i,j}= \frac{1}{3} (\mathbf{W}^{k,l}_{i,j} + 2\mathbf{W}^{k,*}_{i,j}) + \frac{1}{3} \hat{\tau} \mathbf{L} \mathbf{W}^{k,l}_{i,j} 	\\[2mm]
				\hspace{4em}+ \frac{1}{3} \hat{\tau} \mathbf{L} \mathbf{W}^{k,*}_{i,j} + \frac{1}{6} \hat{\tau}^2 \mathbf{L}^2 \mathbf{W}^{k,*}_{i,j}.
			\end{array}
			\right.
			\label{rk}
		\end{equation}
		Here, \(\mathbf{W}^{k,l}_{i,j}\) represents the value of \(\mathbf{W}^k\) at the spatial coordinates \((x_i, y_j)\) and the temporal coordinate \(\hat{\tau}_{l}\), i.e., \(\mathbf{W}^{k,l}_{i,j} = \mathbf{W}^k(x_i, y_j, \hat{\tau}_{l})\). Additionally, the differential operator is defined as \(\mathbf{L}^2 = \mathbf{L} \cdot \mathbf{L} = \text{Diag}[D, D]\). The weight parameter $\eta$ plays a crucial role in enhancing numerical performance, as established in previous literature~\cite{ma2011nearly}. In our current framework, Fig.~\ref{fig:tau_vs_eta} (provided in the Appendix) depicts the relationship between $\eta$ and the optimal iteration step size $\hat{\tau}_{\text{max}}$, while Figure~\ref{fig:tau_vs_b_eta} outlines the corresponding optimal parameter selection strategy for $\eta$ in conjunction with parameter $b$.
	\end{enumerate}
	
	This numerical scheme satisfies all conditions of the Lax equivalence theorem: the wave equation constitutes a well-posed linear problem, the discretization is consistent due to spectral accuracy in space and fourth-order accuracy in time, and stability is guaranteed under the condition derived via Fourier analysis (see Eq.~\eqref{eq:stability_condition}). Consequently, the numerical solution converges to the exact solution of the wave equation.

	\subsection{Summary of Algorithm Workflow}
	The proposed approach is summarized in Algorithm \ref{alg1} as an iterative algorithm.
	\begin{algorithm}[!t]
		\caption{Proposed Approach in an Iterative Algorithmic Form}\label{alg:alg1}
		\begin{enumerate}
			\item Initialize iteration count, $k = 0$.
			\item Using SDF initialize $\phi^0$ by $C^0$.
			\item For image segmentation tasks with varying characteristics and requirements, an appropriate model is selected to compute the initial velocity field $v_0^k$.
			\item To compute $\phi^{k+1}$, PDEs given by Eq. \eqref{hmcf_apply} is solved using the weighted fourth-order R-K algorithm.
			\item Reinitialize $\phi^{k+1}$ to the SDF to the curve (this step is optional).
			\item Check for convergence.
			\begin{itemize}
				\item If not, then update $k=k+1$, recall steps 3-5.
				\item If converged, stop the curve evolution.
			\end{itemize}
		\end{enumerate}
		\label{alg1}
	\end{algorithm}
	
    	\begin{figure}[!t]
    		\centering
    		\includegraphics[width=0.48\textwidth]{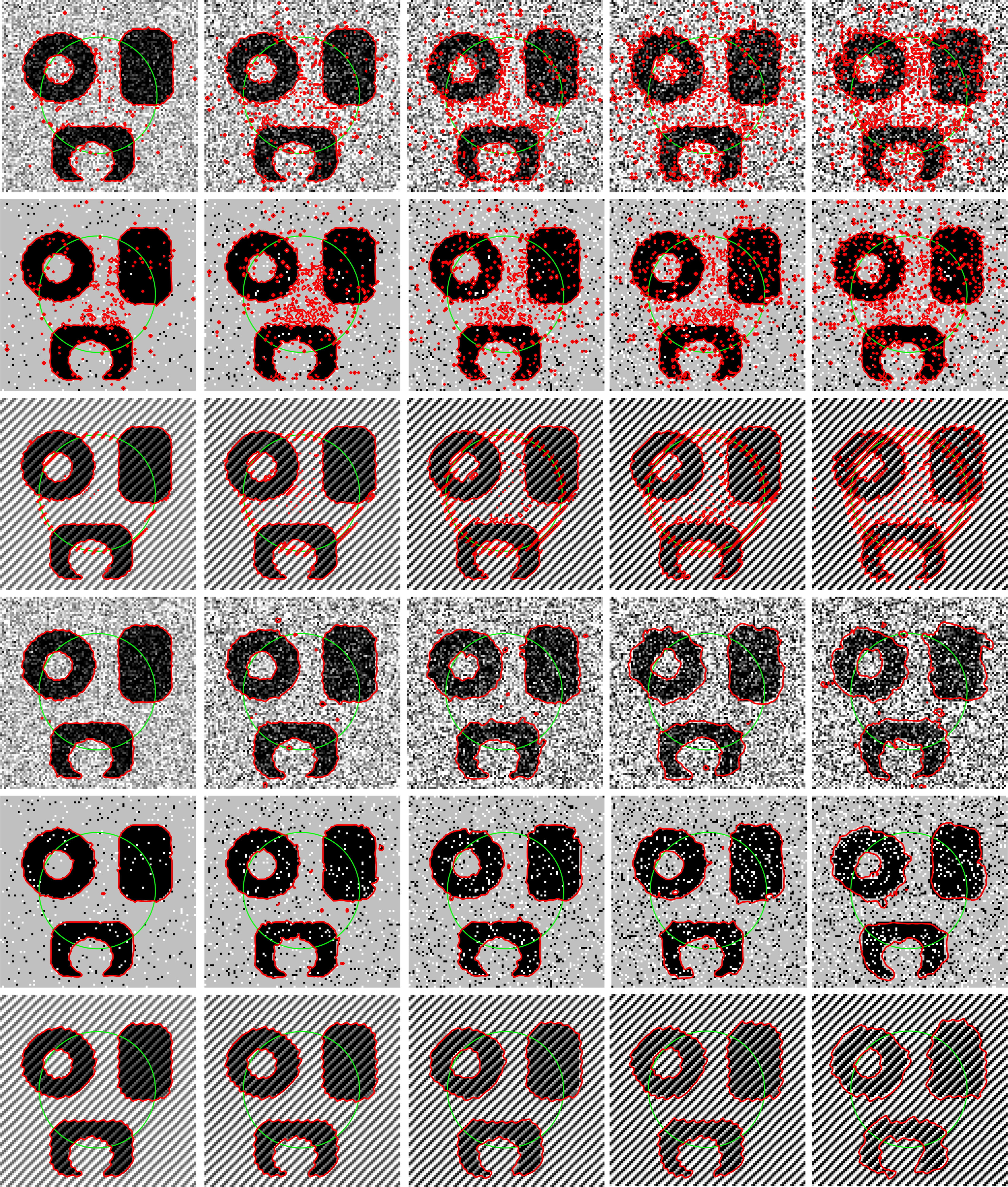}
             \caption{%
            Segmentation results (red curves) and initial contours (green curves) for C-V (rows 1--3) and HMCF-C-V (rows 4--6) under multi-type noise. %
            Left-to-right blocks: Gaussian (0.05--0.25), salt\&pepper (0.05--0.25) and periodic (100--255, frequency 0.2) noise levels.%
            }
    		\label{fig:sensitivity_result}
    	\end{figure}
    	\begin{figure}[!t]
    		\centering
    		\includegraphics[width=0.5\textwidth]{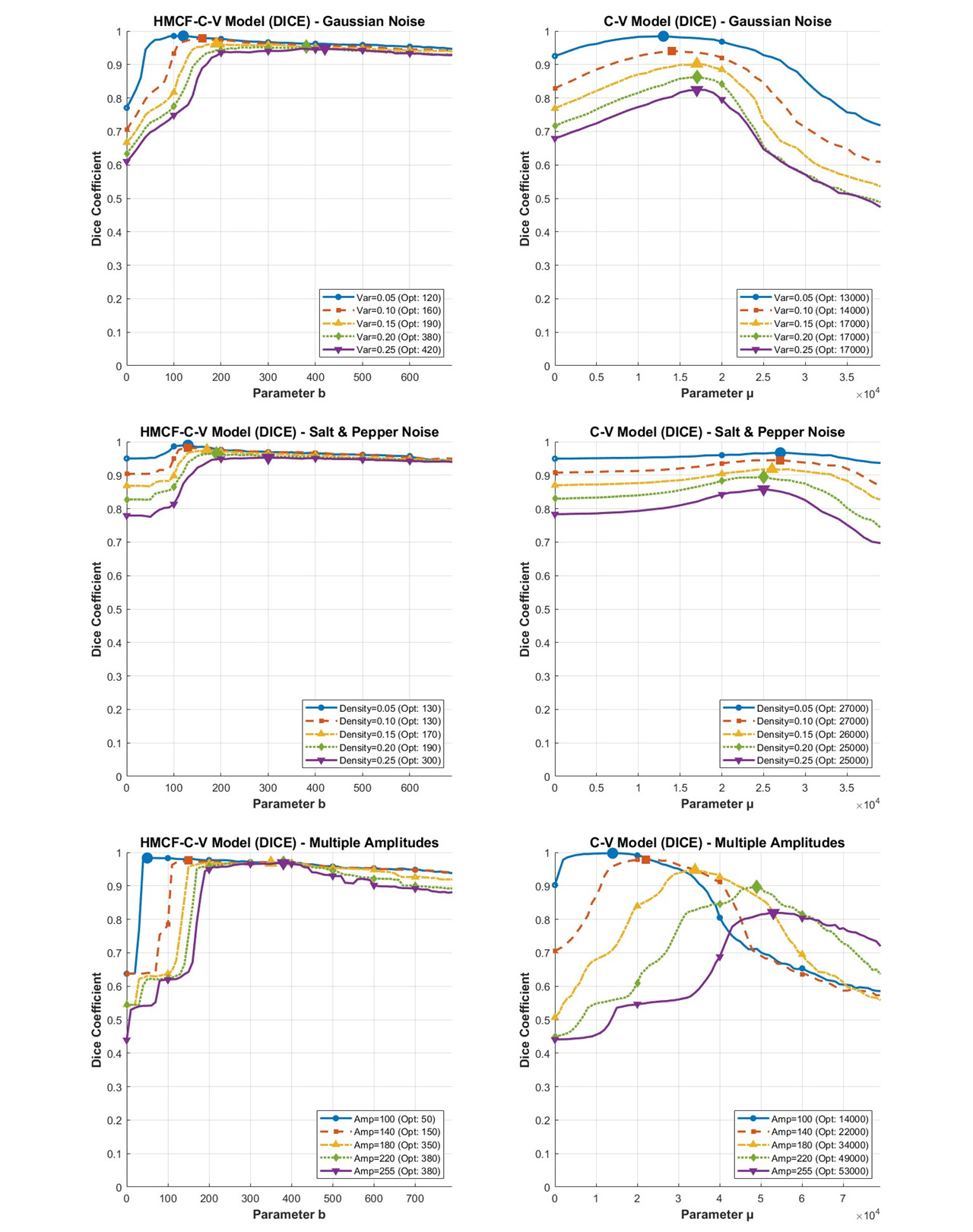}
    		\caption{
    			Variation of Dice metric with curvature parameters ($b$ for HMCF-C-V, $\mu$ for C-V) under different noise conditions. Left column: HMCF-C-V model; Right column: C-V model. Rows 1-3 correspond to Gaussian, salt\&pepper, and periodic noise with increasing intensities, respectively.
    		}
    		\label{line:sensitivity_line_dice}
    	\end{figure}
    \begin{figure}[http]
    		\centering
    		\includegraphics[width=0.5\textwidth]{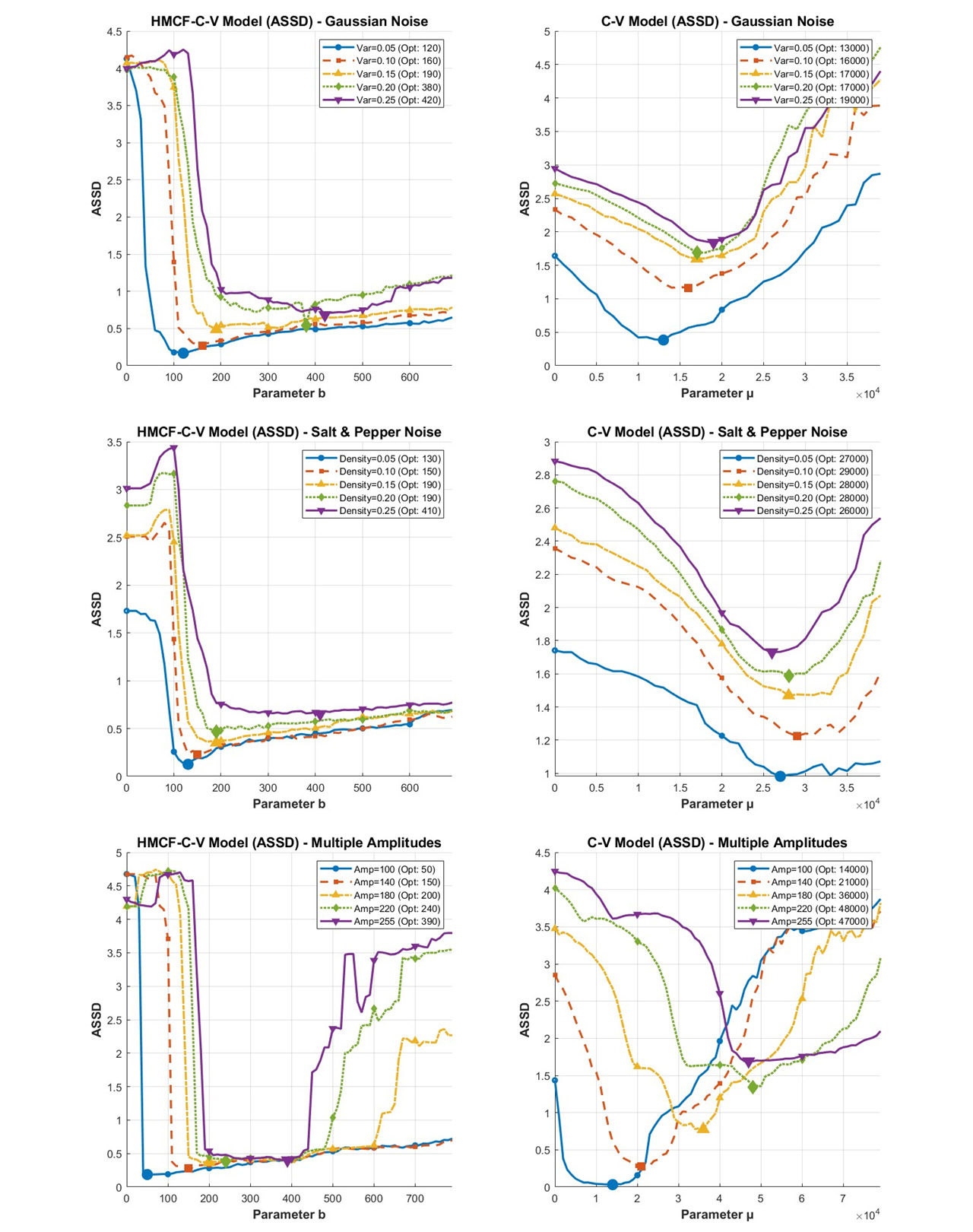}
    		\caption{
    			Variation of ASSD metric with curvature parameters ($b$ for HMCF-C-V, $\mu$ for C-V) under different noise conditions. Left column: HMCF-C-V model; Right column: C-V model. Rows 1-3 correspond to Gaussian, salt\&pepper, and periodic noise with increasing intensities, respectively.
    		}
    		\label{line:sensitivity_line_assd}
    	\end{figure}
	\section{EXPERIMENTS AND RESULTS}\label{sec:E}
	We conclude this paper by presenting comprehensive numerical results by applying our model for a diverse set of synthetic and real images, encompassing various velocity fields, contour configurations, and geometric shapes. In our numerical experiments, the parameters are generally chosen as follows: the time step \(\tau=0.1\), and the weight parameter \(\eta\)=0.7. For the Heaviside functions, we exclusively employ the approximations $H_{\varepsilon}$ with $\varepsilon = 1$.

	Among the parameters, the curvature coefficient \(b\), which serves as a scaling factor, varies across experiments. It is set to small values for images with low-level noise or when the goal is to extract as many targets as possible, and to large values for images dominated by high-level noise, tasks demanding very smooth boundaries, or larger grouped structures.
	
 	All computations were conducted on a workstation equipped with an Intel(R) Core(TM) i5-8265U CPU @ 1.60GHz and 8 GB RAM.

    In Secs.~\ref{subsec:Parameter},~\ref{subsec:Noise}~and~\ref{subsec:Efficiency}, the velocity fields of both the HMCF and PMCF frameworks are designed to exclude distance penalty terms and employ identical reinitialization strategies. This configuration ensures that the experimental outcomes purely reflect the distinctions between HMCF and PMCF—specifically, their respective smoothing mechanisms driven by curvature-based evolution, rather than effects attributable to external regularization terms. In Sec.~\ref{subsec:Accuracy}, distance-penalty terms are retained to allow each model to reach its best possible accuracy on the benchmark dataset.
    
  \begin{table}[!t]
    \centering
    \caption{Optimal Curvature Parameters for HMCF-C-V ($b$) and C-V ($\mu$) Models Under Different Noise Conditions}
    \label{tab:optimal_parameters}
    \small
    \setlength{\tabcolsep}{4pt}
    \begin{tabular}{@{}lccccc@{}}
    \toprule
    \multirow{2}{*}{Noise Type} & \multirow{2}{*}{Intensity} & \multicolumn{2}{c}{Dice-Optimized} & \multicolumn{2}{c}{ASSD-Optimized} \\
    \cmidrule(lr){3-4} \cmidrule(lr){5-6}
    & & $b$ & $\mu$ & $b$ & $\mu$ \\
    \midrule
    \multirow{5}{*}{Gaussian}
    & 0.05 & 120 & 14,000 & 120 & 13,000 \\
    & 0.10 & 160 & 16,000 & 160 & 16,000 \\
    & 0.15 & 190 & 17,000 & 190 & 17,000 \\
    & 0.20 & 380 & 17,000 & 380 & 17,000 \\
    & 0.25 & 420 & 17,000 & 420 & 19,000 \\
    \midrule
    \multirow{5}{*}{Salt\&Pepper}
    & 0.05 & 130 & 27,000 & 130 & 27,000 \\
    & 0.10 & 130 & 27,000 & 150 & 29,000 \\
    & 0.15 & 170 & 26,000 & 190 & 28,000 \\
    & 0.20 & 190 & 26,000 & 190 & 28,000 \\
    & 0.25 & 300 & 25,000 & 410 & 26,000 \\
    \midrule
    \multirow{5}{*}{Periodic}
    & 100 & 50 & 14,000 & 50 & 14,000 \\
    & 140 & 150 & 22,000 & 150 & 21,000 \\
    & 180 & 350 & 34,000 & 200 & 36,000 \\
    & 220 & 380 & 49,000 & 240 & 48,000 \\
    & 255 & 380 & 53,000 & 390 & 47,000 \\
    \bottomrule
    \end{tabular}
    
    \medskip 
    \parbox{\linewidth}{ 
    \footnotesize
    Note: Optimal curvature parameters $b$ (for HMCF-C-V) and $\mu$ (for C-V) 
    corresponding to the best Dice scores and ASSD values under various noise conditions.
    }
\end{table}
	\subsection{Parameter Sensitivity Analysis}\label{subsec:Parameter}
    
    \begin{table*}[!t]
    \centering
    \caption{Performance Comparison of HMCF-ACMs and PMCF-ACMs under Different Noise Conditions}
    \label{tab:comprehensive_comparison}
    \small
    \setlength{\tabcolsep}{3pt}
    \begin{tabular}{@{}lcccccccccccccccc@{}}
    \toprule
    \multirow{3}{*}{Metric} & \multirow{3}{*}{Model} & \multicolumn{5}{c}{Gaussian Noise} & \multicolumn{5}{c}{Salt\&Pepper Noise} & \multicolumn{5}{c}{Periodic Noise} \\
    \cmidrule(lr){3-7} \cmidrule(lr){8-12} \cmidrule(lr){13-17}
     & & 0.05 & 0.10 & 0.15 & 0.20 & 0.25 & 0.05 & 0.10 & 0.15 & 0.20 & 0.25 & 100 & 140 & 180 & 220 & 255 \\
    \midrule
    \multirow{8}{*}{Dice} 
    & HMCF-C-V & 0.99 & 0.97 & 0.96 & 0.95 & 0.94 & 0.99 & 0.98 & 0.97 & 0.96 & 0.95 & 0.98 & 0.98 & 0.97 & 0.97 & 0.97 \\
    & C-V & 0.98 & 0.94 & 0.90 & 0.86 & 0.82 & 0.97 & 0.94 & 0.92 & 0.88 & 0.87 & 1.00 & 0.98 & 0.95 & 0.90 & 0.82 \\
    & HMCF-LBP & 0.99 & 0.97 & 0.97 & 0.96 & 0.95 & 0.99 & 0.99 & 0.98 & 0.97 & 0.96 & 0.99 & 0.98 & 0.97 & 0.97 & 0.97 \\
    & LBF & 0.99 & 0.95 & 0.92 & 0.89 & 0.84 & 0.99 & 0.94 & 0.92 & 0.89 & 0.89 & 0.99 & 0.98 & 0.97 & 0.91 & 0.83 \\
    & HMCF-LIF & 0.99 & 0.99 & 0.98 & 0.97 & 0.95 & 1.00 & 0.98 & 0.98 & 0.97 & 0.96 & 0.99 & 0.99 & 0.98 & 0.97 & 0.98 \\
    & LIF & 0.98 & 0.94 & 0.90 & 0.89 & 0.83 & 0.98 & 0.95 & 0.93 & 0.90 & 0.85 & 1.00 & 0.98 & 0.96 & 0.92 & 0.83 \\
    & HMCF-FRAGL & 1.00 & 0.99 & 0.98 & 0.97 & 0.96 & 1.00 & 0.99 & 0.98 & 0.97 & 0.97 & 0.99 & 0.99 & 0.98 & 0.98 & 0.97 \\
    & FRAGL & 0.99 & 0.94 & 0.89 & 0.88 & 0.85 & 0.99 & 0.95 & 0.93 & 0.91 & 0.89 & 0.99 & 0.98 & 0.98 & 0.96 & 0.85 \\
    \midrule
    \multirow{8}{*}{ASSD} 
    & HMCF-C-V & 0.15 & 0.43 & 0.57 & 0.69 & 0.78 & 0.17 & 0.24 & 0.34 & 0.47 & 0.63 & 0.19 & 0.28 & 0.37 & 0.39 & 0.40 \\
    & C-V & 0.51 & 1.32 & 1.63 & 1.71 & 1.94 & 0.94 & 1.21 & 1.43 & 1.61 & 1.62 & 0.03 & 0.28 & 0.78 & 1.34 & 1.69 \\
    & HMCF-LBP & 0.14 & 0.15 & 0.45 & 0.66 & 0.75 & 0.10 & 0.27 & 0.27 & 0.38 & 0.56 & 0.18 & 0.29 & 0.31 & 0.35 & 0.36 \\
    & LBF & 0.42 & 1.54 & 1.37 & 1.56 & 1.85 & 0.87 & 1.07 & 1.30 & 1.44 & 1.57 & 0.02 & 0.24 & 0.56 & 1.24 & 1.57 \\
    & HMCF-LIF & 0.13 & 0.14 & 0.41 & 0.56 & 0.66 & 0.14 & 0.17 & 0.26 & 0.22 & 0.47 & 0.17 & 0.25 & 0.36 & 0.36 & 0.39 \\
    & LIF & 0.42 & 1.05 & 1.29 & 1.37 & 1.79 & 0.86 & 1.12 & 1.12 & 1.27 & 1.52 & 0.02 & 0.22 & 0.47 & 1.36 & 1.25 \\
    & HMCF-FRAGL & 0.08 & 0.09 & 0.13 & 0.35 & 0.47 & 0.06 & 0.12 & 0.17 & 0.23 & 0.33 & 0.16 & 0.29 & 0.31 & 0.35 & 0.36 \\
    & FRAGL & 0.21 & 0.82 & 1.12 & 1.35 & 1.53 & 0.67 & 0.97 & 1.09 & 1.26 & 1.43 & 0.02 & 0.20 & 0.33 & 0.89 & 1.03 \\
    \bottomrule
    \end{tabular}
    
    \vspace{0.2cm}
    \footnotesize
    Note: Performance comparison of HMCF-based models and traditional models under Gaussian, salt\&pepper, and periodic noise conditions. 
    Higher Dice values and lower ASSD values indicate better segmentation performance.
    \end{table*}
    \begin{figure*}[http]
    \centering
    \includegraphics[width=1\textwidth]{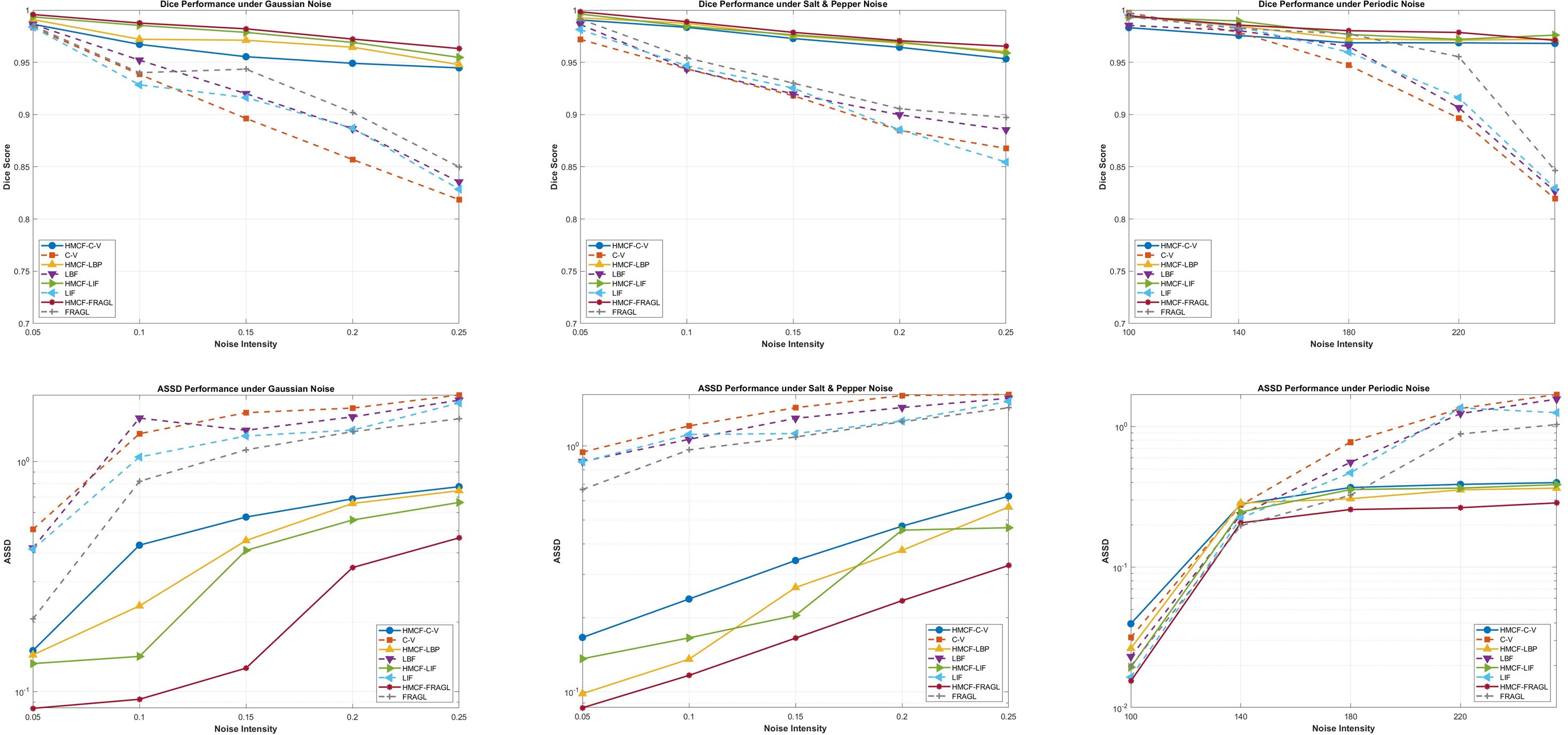}
    \caption{Segmentation performance comparison between HMCF-ACMs and PMCF-ACMs under different noise conditions as reported in Table~\ref{tab:comprehensive_comparison}. Row 1: Dice coefficients; Row 2: ASSD values. From left to right in each row: Gaussian noise, salt\&pepper noise, and periodic noise.}
    \label{fig:noise_dice}
\end{figure*}
        In this section, we evaluate the robustness of the proposed method under various noise conditions using a synthetic image. A comprehensive parameter sensitivity analysis is conducted by comparing segmentation results with the corresponding ground truth, utilizing the Dice Similarity Coefficient (Dice) and the Average Symmetric Surface Distance (ASSD) as quantitative metrics.
        
        The Dice coefficient measures the spatial overlap between the segmentation result \(S\) and the ground truth \(G\), defined as
        \begin{equation}
        \text{Dice}(S, G) = \frac{2|S \cap G|}{|S| + |G|}.
        \end{equation}
        
        The ASSD metric quantifies the average surface distance between the segmentation boundary and the ground truth, formulated as
        \begin{equation}
        \text{ASSD}(S, G) = \frac{1}{|S| + |G|} \left( \sum_{s \in S} d(s, G) + \sum_{g \in G} d(g, S) \right),
        \end{equation}
        where \(d(s, G)\) and \(d(g, S)\) denote the minimum Euclidean distances from point \(s\) (or \(g\)) to the boundary of region \(G\) (or \(S\)).
        
        Fig.~\ref{fig:sensitivity_result} presents a visual comparison of segmentation results under different noise levels, with both models configured using their optimally tuned curvature parameters. It is observed that as the noise intensity increases, the segmentation performance of the C-V model deteriorates significantly, whereas the proposed HMCF-C-V model maintains stable and accurate segmentation results.
        
        Quantitative evaluations shown in Figs.~\ref{line:sensitivity_line_dice} and~\ref{line:sensitivity_line_assd} further highlight the differences in parameter sensitivity. To achieve acceptable segmentation performance—characterized by high Dice coefficients or low ASSD values—the HMCF-C-V model requires a curvature parameter \(b\) within a relatively narrow range of \([0, 1000]\) across Gaussian, salt\&pepper, and periodic noise conditions. In contrast, the C-V model demands a much wider parameter range for \(\mu\), typically within \([0, 80{,}000]\), indicating a significantly higher tuning burden.
        
        Table~\ref{tab:optimal_parameters} summarizes the optimal parameter values corresponding to the best Dice scores and minimal ASSD values across all noise types and intensity levels. The HMCF-C-V model consistently achieves superior performance with \(b \in [50, 420]\), while the C-V model requires \(\mu \in [13{,}000, 53{,}000]\). These results demonstrate that the proposed HMCF-C-V model exhibits enhanced parameter stability and is more amenable to practical deployment with minimal manual tuning.

        \begin{figure*}[!t]
    		\centering
    		\includegraphics[width=1\textwidth]{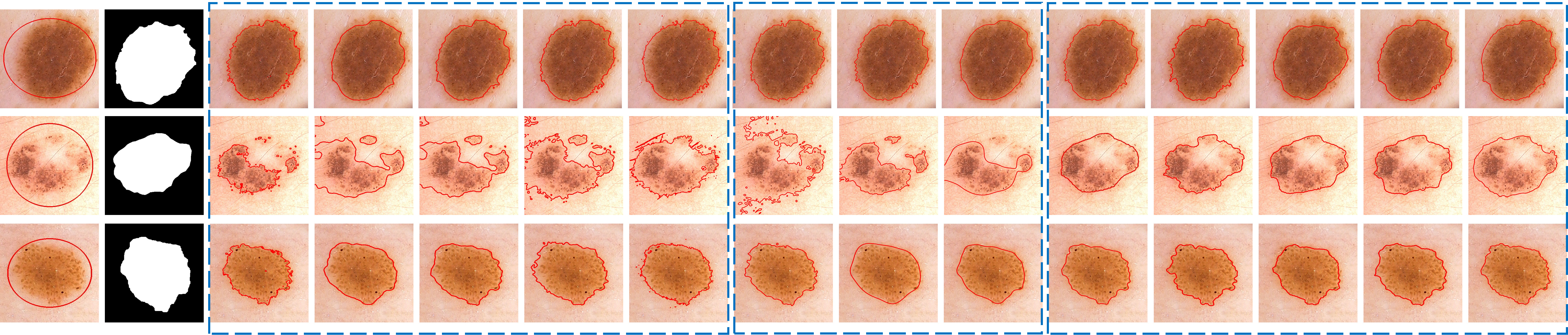}
    		\caption{Comparative segmentation results on sample images from the PH\textsuperscript{2} database~\cite{mendonca2013ph2}. Column 1: input image; Column 2: ground truth; Columns 3--7: C-V, LBF, LIF, LOG\&RSF, FRAGL; Columns 8--10: U-Net, Mask R-CNN, DeepLabV3+; Columns 11--15: HMCF-C-V, HMCF-LBF, HMCF-LIF, HMCF-LOG\&RSF, HMCF-FRAGL.}
    		\label{fig:accuracy_result}
    	\end{figure*}

    \subsection{Noise Sensitivity Analysis}\label{subsec:Noise}
        Visual inspection of Fig.~\ref{fig:noise_dice}, along with the extremal values of the sensitivity curves shown in Figs.~\ref{line:sensitivity_line_dice}--\ref{line:sensitivity_line_assd}, confirms that the HMCF--C--V model consistently outperforms the conventional C--V model under all noise conditions considered.

        To provide a more comprehensive robustness assessment, we extend the evaluation to the full HMCF-ACM family—including HMCF--C--V, HMCF--LBF, HMCF--LIF, and HMCF--FRAGL—and compare them against their PMCF-based counterparts (C--V \cite{chan2001active}, LBF \cite{li2011level}, LIF \cite{zhang2010active}, and FRAGL \cite{FRAGL}). The overall performance is summarized in Table~\ref{tab:comprehensive_comparison}, with the corresponding trends displayed in Fig.~\ref{fig:noise_dice}.
        
        Collectively, the results demonstrate that HMCF--ACMs achieve clearly higher Dice scores and lower ASSD values than PMCF--ACMs when segmenting synthetic images corrupted by various types and intensities of noise. Moreover, while the segmentation accuracy of PMCF-based models declines sharply as noise increases, HMCF variants exhibit a much more gradual performance degradation, highlighting their superior noise tolerance and operational stability.

    		\begin{table}[!t]
    		\centering
    		\caption{Performance Comparison of Various Methods in Terms of Accuracy, Sensitivity, and Dice Metrics}
    		\label{tab:method_performance}
    		\begin{tabular}{lccc}
    			\toprule
    			\textbf{Method} & \textbf{Accuracy (\%)} & \textbf{Sensitivity (\%)} & \textbf{Dice (\%)} \\
    			\midrule
    			HMCF-FRAGL & 96.59 & 96.94 & 96.88 \\
    			FRAGL & 96.10 & 95.68 & 96.21 \\
    			HMCF-LoG\&RSF & 96.54 & 93.65 & 94.64 \\
    			LoG\&RSF & 95.23 & 92.45 & 93.49 \\
    			HMCF-LIF & 87.96 & 85.63 & 84.82 \\
    			LIF & 85.19 & 83.32 & 81.39 \\
    			HMCF-LBP & 84.03 & 80.20 & 83.64 \\
    			LBF & 83.21 & 78.23 & 80.82 \\
    			HMCF-C-V & 76.45 & 71.55 & 61.32 \\
    			C-V & 65.35 & 64.65 & 56.12 \\
    			u-net & 83.36 & 85.45 & 81.54 \\
    			Mask R-CNN & 93.70 & 96.90 & 90.40 \\
    			DeeplabV3+ & 92.30 & 94.30 & 89.00 \\
    			\bottomrule
    		\end{tabular}
    		
    		\vspace{0.2cm}
    		\footnotesize
    		Note: All values are percentages. Higher values indicate better performance for each metric.
    	\end{table}
    \begin{figure}[!t]
    	\centering
    	\includegraphics[width=0.49\textwidth]{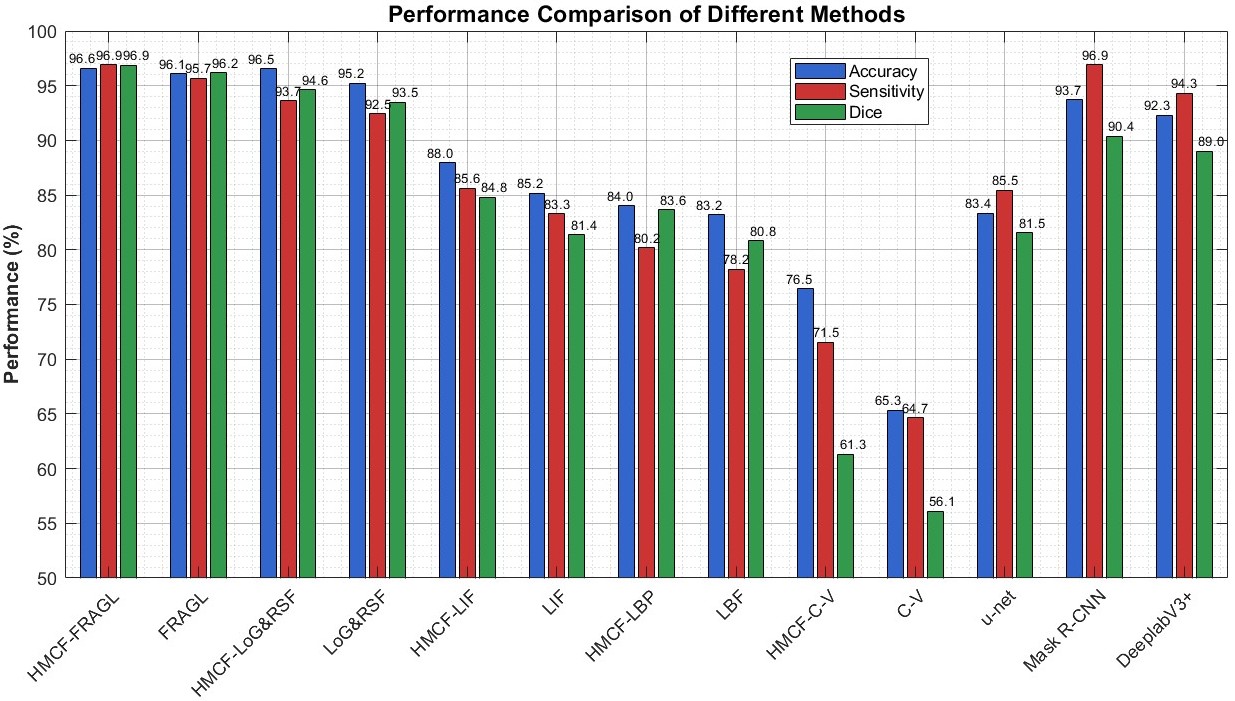}
    	\caption{Segmentation quality comparison of HMCF-ACMs, PMCF-ACMs, and neural networks on the PH\textsuperscript{2} database~\cite{mendonca2013ph2}.}
    	\label{fig:ph2_Barchart}
    \end{figure}
\subsection{Segmentation Accuracy Analysis}\label{subsec:Accuracy}
        This section quantitatively evaluates the segmentation accuracy of the proposed HMCF-ACMs against PMCF-ACMs on dermoscopic skin-lesion images from the PH\(^2\) database~\cite{mendonca2013ph2}.
        All images were resampled to \(512\times512\) pixels and assessed with three standard clinical metrics: segmentation accuracy, sensitivity, and Dice index.
        Accuracy is defined as the ratio of correctly classified pixels,
        \begin{equation}
        \text{Accuracy}=\frac{\mathrm{TP}+\mathrm{TN}}{\mathrm{TP}+\mathrm{TN}+\mathrm{FP}+\mathrm{FN}},
        \end{equation}
        whereas sensitivity emphasises the model's capability to detect true lesion regions,
        \begin{equation}
        \text{Sensitivity}=\frac{\mathrm{TP}}{\mathrm{TP}+\mathrm{FN}}.
        \end{equation}
        
        Fig.~\ref{fig:accuracy_result} visually compares the initial contour, ground-truth annotation, and the segmentation masks produced by PMCF-ACMs, HMCF-ACMs, and representative deep models.
        HMCF-ACMs consistently yield smoother lesion boundaries that adhere more closely to the expert delineations.
        
        Table~\ref{tab:method_performance} summarises the quantitative results for HMCF-FRAGL, FRAGL~\cite{FRAGL}, HMCF-LoG\&RSF, LoG\&RSF~\cite{niaz2023edge}, HMCF-LIF, LIF~\cite{zhang2010active}, HMCF-LBF, LBF~\cite{li2011level}, HMCF-C-V, C-V~\cite{chan2001active}, U-Net~\cite{ronneberger2015u}, Mask R-CNN~\cite{8372616}, and DeepLabv3+~\cite{chen2018encoder}.
        Across all three metrics, HMCF-ACMs significantly outperform their PMCF counterparts.
        The bar chart in Fig.~\ref{fig:ph2_Barchart} further highlights the superior segmentation accuracy achieved by HMCF-based active contours, several of which even surpass the evaluated neural-network-based methods.

    \subsection{Computational Efficiency Analysis}\label{subsec:Efficiency}
    	\begin{figure*}[!t]
		\centering
		\includegraphics[width=1\textwidth]{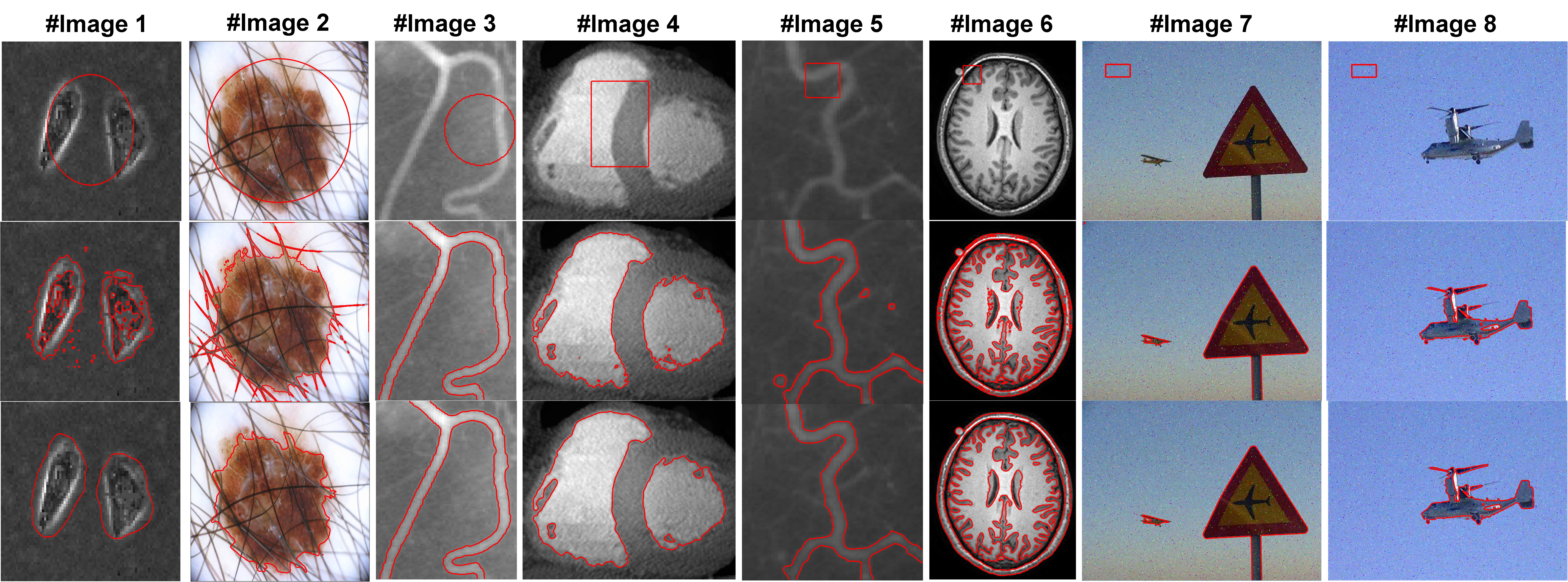}
		\caption{Visualization of segmentation performance. Row 1: Input image with initial contour (red); Row 2: Segmentation results using baseline PMCF-ACMs; Row 3: Segmentation results using the proposed HMCF-ACMs.}
		\label{speed}
	\end{figure*}
	\begin{table}[!t]
		\centering
		\caption{Cpu Times for Contour Evolution over the Images in Fig.~\ref{speed}}
		\label{tab:speed_comparison}
		\begin{tabular}{llrrl}
			\toprule
			\textbf{Image} & \textbf{Model} & \textbf{Time (s)} & \textbf{Iterations} & \textbf{Parameter} \\
			\midrule
			\multirow{2}{*}{Image 1} & HMCF-C-V & 0.34 & 15 & $ b = 2200 $ \\
			& C-V       & 0.21 & 15 & $ \mu = 2200 $ \\
			\addlinespace
			\multirow{2}{*}{Image 2} & HMCF-C-V & 14.13 & 15 & $ b = 8000 $ \\
			& C-V       & 13.59 & 15 & $ \mu = 8000 $ \\
			\addlinespace
			\multirow{2}{*}{Image 3} & HMCF-LBF & 4.50 & 95 & $ b = 30 $ \\
			& LBF       & 4.48 & 95 & $ \mu = 30 $ \\
			\addlinespace
			\multirow{2}{*}{Image 4} & HMCF-LBF & 2.96 & 30 & $ b = 200 $ \\
			& LBF       & 2.26 & 30 & $ \mu = 200$ \\
			\addlinespace
			\multirow{2}{*}{Image 5} & HMCF-LPF & 3.98 & 55 & $ b = 200$ \\
			& LPF       & 2.52 & 55 & $ \mu = 200$ \\
			\addlinespace
			\multirow{2}{*}{Image 6} & HMCF-LPF & 6.31 & 31 & $ b = 200$ \\
			& LPF       & 6.11 & 31 & $ \mu = 200 $ \\
			\addlinespace
			\multirow{2}{*}{Image 7} & HMCF-FRAGL & 2.90 & 40 & $ b = 1 $ \\
			& FRAGL      & 2.50 & 40 & $ \mu = 1 $ \\
			\addlinespace
			\multirow{2}{*}{Image 8} & HMCF-FRAGL & 2.98 & 40 & $ b = 1 $ \\
			& FRAGL      & 2.53 & 40 & $ \mu = 1 $ \\
			\bottomrule
		\end{tabular}
	\end{table}

    Figure~\ref{speed} illustrates segmentation results on challenging natural images characterized by heterogeneous edges, hair occlusion, uneven illumination, and noise. Row~1 shows the original inputs; Row~2 presents results obtained with baseline PMCF-ACMs (C-V~\cite{chan2001active}, LBF~\cite{li2011level}, LPF~\cite{ding2018active}, FRAGL~\cite{FRAGL}); Row~3 displays the corresponding outcomes of the proposed HMCF-ACMs (HMCF-C-V, HMCF-LBF, HMCF-LPF, HMCF-FRAGL). All models successfully delineated the target regions. For Images 3–8, visual assessment indicates that, under identical curvature parameters, HMCF-ACMs deliver contour-smoothing performance comparable to that of PMCF-ACMs. In contrast, for the more challenging Images 1–2, HMCF-ACMs exhibit markedly superior robustness to extreme imaging conditions.
    	
    	Table~\ref{tab:speed_comparison} presents the CPU times for contour evolution under identical curvature parameters and iteration counts across the images in Fig.~\ref{speed}. The results confirm that the computational efficiency of PMCF-ACMs and HMCF-ACMs is nearly identical, with HMCF-ACMs exhibiting only a marginal increase in processing time—a justifiable trade-off given their enhanced segmentation performance.

\section{Conclusion}
	This paper presents a systematic advancement in ACMs through four interconnected contributions. We first establish a numerically equivalent hyperbolic formulation for the HMCF framework by integrating level set methods with SDF preservation techniques, thereby laying the theoretical foundation for subsequent developments. Building on this foundation, we pioneer the application of HMCF to regulate contour smoothness in image segmentation, overcoming the poor performance of PMCF-ACMs on high-intensity-noise images. The computational framework is boosted by an optimized weighted fourth-order Runge–Kutta scheme paired with DCT-based spatial discretization, maintaining high accuracy while significantly reducing computational cost. Comprehensive experimental validation demonstrates that both HMCF-driven models achieve precise segmentation with exceptional noise immunity, achieved through task-adaptive configuration of velocity fields and initial contours within the unified HMCF paradigm. Leveraging the low parameter sensitivity of HMCF-ACMs, future work will investigate their integration as a learnable shape prior within neural network architectures.

\appendix
\section*{Stability Analysis of Numerical Algorithms}
Consider the discrete cosine wave solution in two-dimensional space
\begin{equation}
	\mathbf{W}_{i,j}^{l} = \mathbf{W}^{l} \cos\left(k_x i \Delta x \right) \cos\left(k_y j \Delta y \right),
	\label{eq:dct_wave}
\end{equation}
where $k_x = \pi \cos \theta_1$ and $k_y = \pi \cos \theta_2$ denote the discrete wavenumbers in the $x$ and $y$ directions, respectively. 
Substitution of this DCT-based wave solution into the numerical scheme yields the amplification equation

\begin{equation}
	\mathbf{W}^{l+1} = G \mathbf{W}^{l},
	\label{eq:amplification}
\end{equation}
where the amplification matrix $G$ is defined as

\begin{equation}
	G = \begin{pmatrix}
		g_{11} & g_{12} \\
		g_{21} & g_{22}
	\end{pmatrix}. 
	\label{eq:amplification_matrix}
\end{equation}
The elements of matrix $G$ are analytically derived as follows
\begin{align*}
	g_{11} &= \frac{2}{3} \beta - \frac{1}{6} \hat{\tau}^2 (\eta + (2 - \eta) \beta) b \pi^2 (\cos^2 \theta_1 + \cos^2 \theta_2) + \frac{1}{3}, \\
	g_{12} &= -\frac{2}{3} \hat{\tau} b \pi^2 (\cos^2 \theta_1 + \cos^2 \theta_2) \\
	&\quad + \frac{1}{12} \hat{\tau}^3 b^2 \pi^4 (\cos^2 \theta_1 + \cos^2 \theta_2)^2, \\
	g_{21} &= \frac{1}{3} \hat{\tau} (1 + \eta + (2 - \eta) \beta) - \frac{1}{12} \hat{\tau}^3 b \pi^2 (\cos^2 \theta_1 + \cos^2 \theta_2), \\
	g_{22} &= 1 - \frac{1}{3} \hat{\tau}^2 b \pi^2 (\cos^2 \theta_1 + \cos^2 \theta_2) \\
	&\quad + \frac{1}{24} \hat{\tau}^4 b^2 \pi^4 (\cos^2 \theta_1 + \cos^2 \theta_2)^2, \\
	\beta &= 1 - \pi^2 (\cos^2 \theta_1 + \cos^2 \theta_2) \left( \hat{\tau} + \frac{1}{4} \hat{\tau}^2 \right).
\end{align*}
	\begin{figure}[!t]
	\centering
	\includegraphics[width=0.45\textwidth]{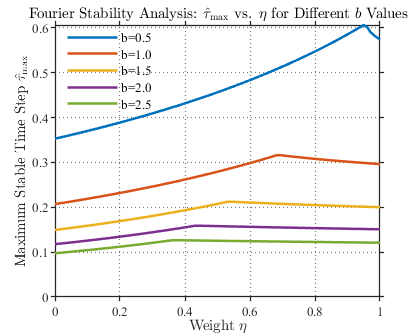}
	\caption{
		Maximum stable time step $\hat{\tau}_{max}$ versus weight parameter $\eta$ for the weighted Runge-Kutta method with varying parameters $b = \{0.5, 1, 1.5, 2, 2.5\}$
	}
	\label{fig:tau_vs_eta}
\end{figure}

\begin{figure}[!t]
	\centering
	\includegraphics[width=0.5\textwidth]{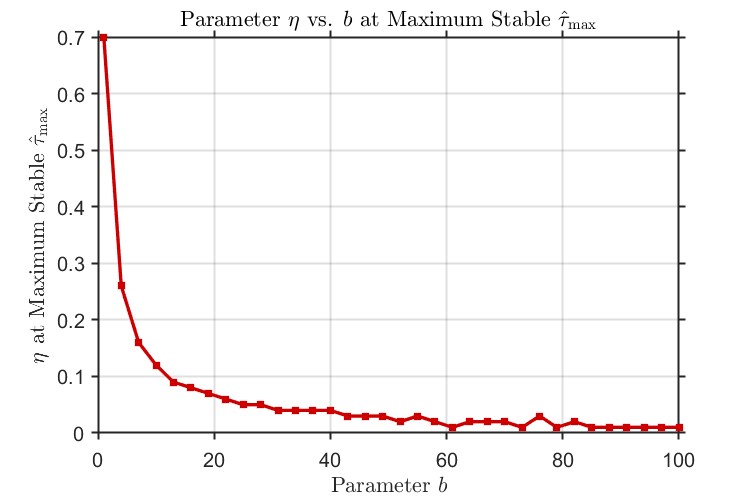}
	\caption{Optimal weight parameter $\eta$ versus parameter $b$ for maximizing stable time step $\hat{\tau}_{max}$.}
	\label{fig:tau_vs_b_eta}
\end{figure}

\begin{figure}[!t]
	\centering
	\includegraphics[width=0.5\textwidth]{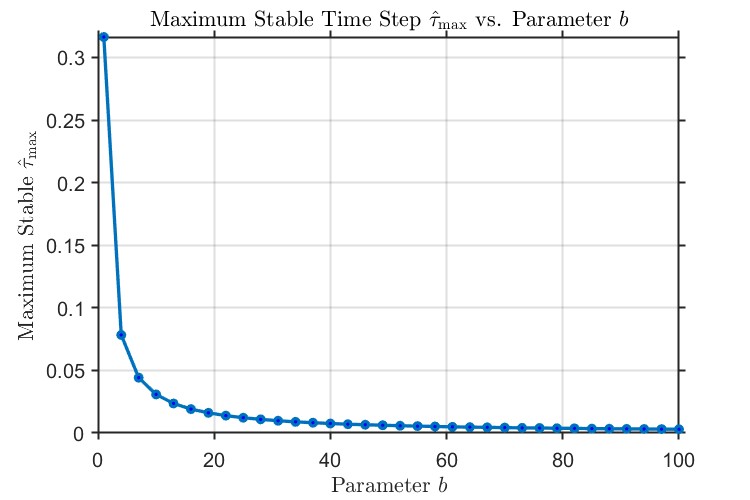}
	\caption{Maximum stable time step $\hat{\tau}_{\mathrm{max}}$ versus parameter $b$.}
	\label{fig:tau_vs_b}
\end{figure}
Numerical solution of the eigenvalue problem $|\lambda(G)| \leq 1$ yields the following stability criterion
\begin{equation}
	\hat{\tau}_{\mathrm{max}} \approx  \frac{0.316}{b},
	\label{eq:stability_condition}
\end{equation}
where $\hat{\tau}_{\mathrm{max}}$ represents the maximum admissible time step that ensures numerical stability of the algorithm. This derived stability condition is empirically validated through comprehensive numerical experiments, as illustrated in Figs. \ref{fig:tau_vs_eta}, \ref{fig:tau_vs_b_eta}, and \ref{fig:tau_vs_b}, which systematically characterize the relationship between the maximum stable time step $\hat{\tau}$ and the critical parameters $\eta$ and $b$.

	\bibliographystyle{IEEEtran}
	\bibliography{ref.bib}

\end{document}